\documentclass{article}

\usepackage{microtype}
\usepackage{graphicx}
\usepackage{subcaption}
\usepackage{booktabs} % for professional tables

\usepackage{hyperref}
\usepackage{multirow}

\usepackage[preprint]{icml2026}

\usepackage{amsmath}
\usepackage{amssymb}
\usepackage{mathtools}
\usepackage{amsthm}

\usepackage[capitalize,noabbrev]{cleveref}

\theoremstyle{plain}

\theoremstyle{definition}

\theoremstyle{remark}

\usepackage[textsize=tiny]{todonotes}

\newcommand{\mname}{ReV}
\usepackage[table]{xcolor}
\usepackage{siunitx}
\usepackage{xspace}
\definecolor{lightblue}{RGB}{202,239,252}

\newcommand{\ie}{\textit{i}.\textit{e}., }
\newcommand{\eg}{\textit{e}.\textit{g}., }
\newcommand{\cf}{\textit{cf}. }

\begin{document}

\twocolumn[
  \icmltitle{Referring-Aware Visuomotor Policy Learning for Closed-Loop Manipulation}



  \icmlsetsymbol{equal}{*}

  \begin{icmlauthorlist}
    \icmlauthor{Jiahua Ma}{equal,sysu}
    \icmlauthor{Yiran Qin}{equal,oxford}
    \icmlauthor{Xin Wen}{equal,sysu}
    \icmlauthor{Yixiong Li}{sysu}
    \icmlauthor{Yuyu Sun}{sysu}
    \icmlauthor{Yulan Guo}{sysu}
    \icmlauthor{Liang Lin}{sysu}
    \icmlauthor{Ruimao Zhang}{sysu}
  \end{icmlauthorlist}

  \icmlaffiliation{sysu}{Sun Yat-sen University}
  \icmlaffiliation{oxford}{Oxford University}
  \icmlcorrespondingauthor{Ruimao Zhang}{zhangrm27@mail.sysu.edu.cn}

  \icmlkeywords{Robotic Manipulation, Diffusion Policy}

  \vskip 0.3in  
]



\printAffiliationsAndNotice{\icmlEqualContribution}

\begin{figure*}[t]
    \centering
    \setlength{\fboxrule}{0pt}
    \framebox{{\includegraphics[width=\linewidth]{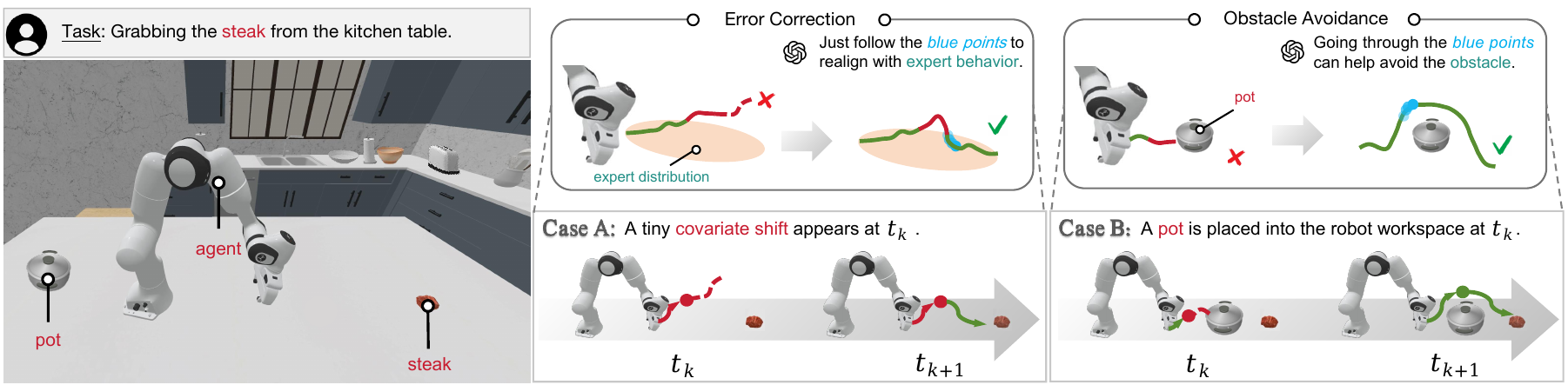}}}
    \caption{The role of the proposed Referring-Aware Visuomotor Policy (\mname{}) in mitigating out-of-distribution failures. 
   In executing the task \textit{``Grabbing the steak from the kitchen table"}, several challenges can be encountered.   
   \textbf{Case A}: a small covariate shift at $t_k$ quickly leads to a compounding misalignment with expert demonstrations, driving the robot into unseen states and resulting in task failure.
   \textbf{Case B}: an unforeseen obstacle (\eg a pot) emerges at $t_k$ to dynamically block the robot's path, leading to failure as well. 
   Unlike traditional imitation learning methods that often struggle to generalize to unseen states or observations, the proposed \mname{} employs coupled diffusion heads to online react to external sparse referring provided by a human or a high-level reasoning planner. Without requiring additional training data or complex fine-tuning in post-processing, the model can effectively address Case A and Case B in real-world applications.}
    \label{fig:motivation}
\end{figure*}

\begin{abstract}
  This paper addresses a fundamental problem of visuomotor policy learning for robotic manipulation: how to enhance robustness in out-of-distribution execution errors or dynamically re-routing trajectories, where the model relies solely on the original expert demonstrations for training. We introduce the \textbf{Re}ferring-Aware \textbf{V}isuomotor Policy (\mname{}), a closed-loop framework that can adapt to unforeseen circumstances by instantly incorporating sparse referring points provided by a human or a high-level reasoning planner. Specifically, \mname{} leverages the coupled diffusion heads to preserve standard task execution patterns while seamlessly integrating sparse referring via a trajectory-steering strategy. Upon receiving a specific referring point, the global diffusion head firstly generates a sequence of globally consistent yet temporally sparse action anchors, while identifies the precise temporal position for the referring point within this sequence. Subsequently, the local diffusion head adaptively interpolates adjacent anchors based on the current temporal position for specific tasks. This closed-loop process repeats at every execution step, enabling real-time trajectory replanning in response to dynamic changes in the scene. In practice, rather than relying on elaborate annotations, \mname{} is trained only by applying targeted perturbations to expert demonstrations. Without any additional data or fine-tuning scheme, \mname{} achieve higher success rates across challenging simulated and real-world tasks. Project page: \url{https://gaavama.github.io/ReV/}.
\end{abstract}

\section{Introduction}
\label{sec:intro}
With the development of large-scale simulated and real-world robotic datasets~\cite{10611477, nasiriany2024robocasalargescalesimulationeveryday, pmlr-v229-walke23a, chen2025metafoldlanguageguidedmulticategorygarment}, imitation learning based visuomotor policy models~\cite{xue2025demogensyntheticdemonstrationgeneration, singh2025dextrahrgbvisuomotorpoliciesgrasp, 10611491, 9981402, zhang2026touchguide} have shown the ability to perform a wide range of daily tasks. 
% These data-driven visuomotor policies excel at executing instructed tasks but often fail to handle \emph{out-of-distribution} situations due to the lack of generalizable spatial reasoning.
%These data-driven visuomotor policies excel at executing instructed tasks, yet their pure imitation objective provides no explicit mechanism for recovery once the states and observations drifts outside the demonstrated manifold, leaving them vulnerable to \emph{out-of-distribution} situations.
%
However, the imitation objective of these visuomotor policy models provides no explicit mechanism for recovery. 
Consequently, while these data-driven visuomotor policies excel at executing instructed tasks, they become very fragile to \emph{out-of-distribution} (OOD) situations once states and observations drift outside the demonstrated distribution.
As shown in Fig.~\ref{fig:motivation}, they cannot recover from execution errors or replan trajectories to choose safer, more reasonable paths.

%As inllustrated in Fig.～\ref{fig:overview}, they struggle to recover from execution errors or dynamically re-route trajectories to select the safer and more reasonable paths.
% In other words, we desire policies that can actively reason during inference—correcting erroneous trajectories or choosing optimal ones.
% based on external guidance.

To address such an issue, one possible solution is to expand the training distribution with large datasets of errors and human corrections, explicitly training the policy model to recover from errors~\cite{8917583, mandlekar2021matterslearningofflinehuman,mandlekar2023mimicgendatagenerationscalable}. 
However, these approaches require enormous human effort, do not scale well, and may even compromise success rates by introducing suboptimal trajectories.
%
% Another line of work adopts open-loop reasoning planner to handle \emph{out-of-distribution} situations leveraging their rich priors, where a high-level model (\eg a LLM) predicts intermediate robot poses, and rule-based motion or grasping modules~\cite{huang2023voxposercomposable3dvalue}
% (\eg VoxPoser)
% are used to execute them. Although these approaches introduce reasoning into the inference process, their low-frequency rule-based interactions make them brittle in dynamic environments where the scene changes during execution.
%Another line of work employs open-loop reasoning planners~\cite{huang2023voxposercomposable3dvalue} to address \emph{out-of-distribution} scenarios by leveraging the rich prior knowledge embedded in high-level models, such as large language models (LLMs). 
%
Another line of research leverages carefully designed cost or reward functions to guide robots toward collision-free and constraint-satisfying trajectories in unseen scenarios~\cite{janner2022planning, carvalho2025motion, zhao2024dartcontrol}. 
But in dynamic environments, manually specifying such functions becomes impractical—they often fail to generalize to novel constrained settings. 
To address this limitation, \cite{huang2023voxposercomposable3dvalue} introduce reasoning planners that interpret constraints and generate intermediate robot poses by exploiting strong priors from high-level models such as large language models, to generate the trajectory globally.
Nonetheless, these approaches still rely on the rule-based execution modules, whose low-frequency rule-based interactions still make them ill-suited in dynamic environments.
%
% Recent works focus on leveraging open-loop reasoning planners ~\cite{huang2023voxposercomposable3dvalue} to address OOD scenarios in dynamic environments, drawing on the strong priors of high-level models, such as large language models. 
%
% These methods first generate intermediate robot poses, which, however, are then executed by rule-based motion or grasping modules. Although they introduce reasoning into the manipulation, their low-frequency rule-based interactions still make them ill-suited in dynamic environments.
%
Given this, one question naturally arise: \emph{when limited to expert demonstration data for imitation learning, how can we enhance the robustness of the policy model in out-of-distribution situations within dynamic environments?}

%\emph{We desire policies that can handle out-of-distribution situations and adapt to dynamic environments during inference, relying solely on the original training data.}

%These models generate intermediate robot poses, which are subsequently executed through rule-based motion or grasping modules. While such approaches introduce reasoning into the inference process, their low-frequency rule-based interactions make them fragile in dynamic environments where the scene may change during execution.

In this paper, we introduce a referring-awareness visuomotor policy model termed \mname{}, which is a \emph{closed-loop} framework that can incorporate external referring information (\eg from humans or high-level reasoning planners) to enhance both adaptability and generalization. 
By receiving referring information into carefully designed architecture, \mname{} enables the model to flexibly handle error recovery (leading the robot back to the expert distribution) or perform precise goal-oriented adjustments (navigating to safer or more optimal regions), \emph{without requiring} additional training data or complex fine-tuning in post-processing.

% we aim to build a closed-loop visuomotor policy that can incorporate external reasoning information to enhance both adaptability and generalization, without requiring additional training data or iterative fine-tuning. To this end, we propose \mname{}, a closed-loop visuomotor policy driven by external sparse guidance. By injecting reasoning-based external guidance (\eg from humans or reasoning LLMs) into the policy at inference time, \mname{} enables the model to flexibly handle error recovery (guiding the robot back to the expert distribution) and perform precise goal-oriented adjustments (navigating through safer or more optimal regions).

%In practice, \mname{} receives 3D referring points as guidance, rather than open-ended language instructions. These points offer sub-centimeter spatial precision while avoiding the well-known language-following deficits commonly observed in visuomotor policies. 
%
In practice, \mname{} employs a diffusion-based planner with 3D referring-point guidance to generate manipulation trajectories, thereby enabling sub-centimeter spatial precision.
% Compared with the language instructions~\cite{octo_2023}, 
% This approach yields sub-centimeter spatial precision while avoiding the linguistic ambiguities that typically challenge visuomotor policies.
%
% To respond to these points more effectively, we introduce a hierarchical architecture that endows the policy with a global perspective, enabling it to understand and plan with respect to affordance points at the trajectory level. 
Specifically, we introduce a novel architecture for \mname{} that employs coupled diffusion heads to achieve a more effective response to referring points.
% that captures global task execution patterns, enabling trajectory-level planning with respect to these points.
%
%Specifically, we use a transformer-based encoder to assign each referring point a plausible temporal position along the entire trajectory. 
Specifically, upon receiving a referring point, a Temporal-Position Prediction module is firstly adopted to estimate its plausible location along the execution trajectory.
Then the trajectory-steering strategy feeds this temporally-positioned point into the Global Diffusion Head, producing a series of sparse but precise action anchors that reliably reach the specified targets. 
Subsequently, a Local Diffusion Head conducts temporal-dependent interpolation strategies between consecutive anchors, progressively refining them into a smooth and fine-grained trajectory. 
The full model operates recurrently at each inference step, enabling dynamic online replanning in response to evolving scene conditions.

%These temporally-positioned points are then injected into the global diffusion head through our trajectory-steering strategy, generating sparse action anchors that accurately reach the specified points and successfully complete the task. A local diffusion head is then used to learn the temporal-dependent interpolation strategies between these anchors, and finally generate a fine-grained trajectory. The entire model is invoked at every inference step, enabling online replanning as the scene evolves. 

The main \textbf{contributions} can be summarized as follows.
1) We present a referring-aware visuomotor policy that operates within an imitation learning framework. By integrating point-level referring cues with task-specific execution patterns, it enables robots to effectively handle challenging \emph{out-of-distribution} scenarios.
2) We introduce a novel policy model with coupled diffusion heads that generates actions in a coarse-to-fine manner, well supporting closed-loop inference.
3) Extensive experiments in both simulation and real-world settings show that \mname{} outperforms other state-of-the-art visuomotor policies in referring-aware manipulation tasks.

\section{Related Works}
\label{sec:related_work}

\subsection{Visuomotor Policy Models for Manipulation}
Visuomotor policy models integrate visual perception with motor control, enabling robots to perform manipulation tasks in complex and unstructured environments~\cite{shridhar2023perceiver, wang2023mimicplay, ze2023gnfactor, peng2020learning, agarwal2023dexterous, haldar2023teach}. 
% These approaches typically map sensory inputs directly to action commands, supporting end-to-end learning of manipulation behaviors.
%
Two generative paradigms have recently gained significant attention for addressing challenges in this domain.
Autoregressive models~\cite{zhao2023learningfinegrainedbimanualmanipulation, xian2023chaineddiffuser, gong2024carp, cui2022playpolicyconditionalbehavior, shafiullah2022behaviortransformerscloningk, lee2024behaviorgenerationlatentactions, zhang2025autoregressive} decompose the trajectory distribution into a sequence of next-step conditionals. This factorization enables efficient training and fast inference at deployment. 
% And the maturity of transformer-based frameworks further facilitates stable optimization and seamless integration with large-scale pretraining. 
However, this token-by-token generation process lacks the ability to revise earlier decisions, causing small deviations to accumulate and degrade global coherence over time.
Recently, diffusion models~\cite{ho2020denoising, song2020denoising}, 
%
%by learning to denoise complex, high-dimensional sample spaces, 
have proven remarkably effective for trajectory synthesis~\cite{chi2023diffusion, Ze2024DP3, ma2025cdp, su2025dense, wei2024ensuring, wang2025gaudpreinventingmultiagentcollaboration}. Their ability to model intricate, multimodal distributions lets them generate more accurate and flexible robot motions.
However, these methods struggle to handle OOD situations because of their reliance on imitation learning.
% , yielding state-of-the-art success rates on manipulation tasks. 
% Nevertheless, existing works still model temporal dependencies in a predominantly local manner.
%
% In contrast, our method explicitly models long-range dependencies across the entire trajectory through a hierarchical framework, ensuring that the generated motions remain temporally consistent and ultimately guaranteeing a high success rate for manipulation tasks.

\subsection{Robotic Motion Planning}
Traditional motion planning approaches are broadly categorized into sampling-based and optimization-based planning. Sampling-based planners~\cite{karaman2011sampling, gammell2014informed, gammell2020batch, strub2020adaptively} explore feasible trajectories by randomly sampling and connecting collision-free nodes in the configuration space, typically producing trajectories composed of straight-line segments, which lack smoothness. In contrast, optimization-based planners~\cite{urain2022learning, petrovic2022mixtures, le2023accelerating} formulate planning as a numerical optimization problem, directly solving for trajectories that minimize objective functions incorporating collision and smoothness costs. However, their performance heavily relies on the initial planning priors and is prone to local optima.
%
% To overcome the limitations of traditional optimization methods, learning planning priors from data has emerged as a promising direction. 
Recently, diffusion models with their strong multimodal generative capability, have been introduced into robot motion planning. MPD~\cite{carvalho2025motion} employed a diffusion model as a trajectory prior and utilized classifier guidance to incorporate collision costs during inference for trajectory refinement. Similarly, EDMP~\cite{saha2024edmp} adopted an ensemble of cost functions to enhance robustness. 
% These works demonstrate the potential of diffusion models in capturing multimodal trajectory distributions.
%
Nevertheless, existing diffusion-based planning methods generally rely on predefined and fixed reward or loss functions to guide the generation process. This reveals significant limitations when dealing with dynamically changing environments or OOD scenarios: handcrafted, static reward functions struggle to accurately and flexibly encode the full spectrum of constraints and high-level semantics involved in real-world manipulation tasks
% —such as contact mechanics, dynamic obstacle avoidance, and task-completion preferences—
resulting in limited adaptability and generalization in rapidly evolving settings.

\begin{figure*}[t]
    \centering
    \setlength{\fboxrule}{0pt}
    \framebox{{\includegraphics[width=\linewidth]{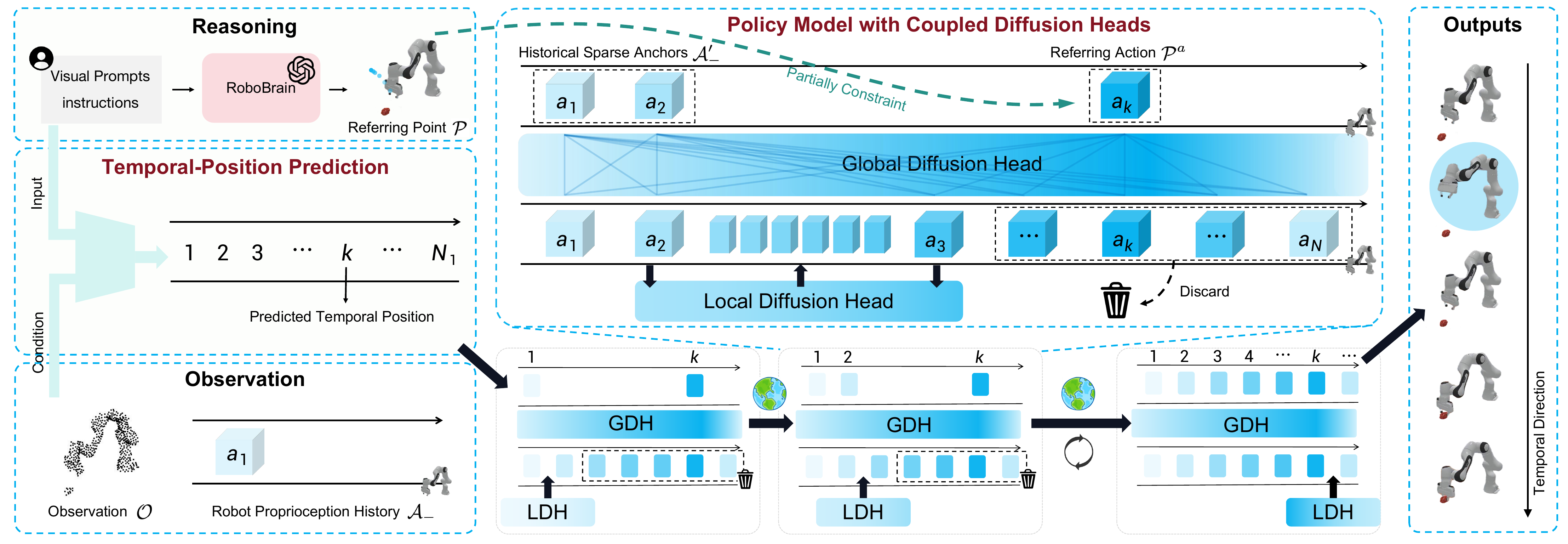}}}
    \caption{\textbf{Overview of our Referring-Aware Visuomotor Policy.} Given the observation $\mathcal{O}$ and robot proprioception history $\mathcal{A}_{-}$, the \emph{Temporal-Position Prediction Module} on the left assign the optimal temporal position $k$ for the referring point $\mathcal{P}$.
    This temporally-positioned referring point $\mathcal{P}$ is then transformed into a temporally-positioned partially-constrained referring action  $\mathcal{P}^{a}$ (detailed in Sec.~\ref{sec:trajectory-steering}) and injected into the right \emph{Policy Model with Coupled Diffusion Heads} to generate the final trajectory $\mathcal{A}$ satisfying the \emph{Manipulation Steering Choreography}. Specifically, at step $i$, the \texttt{GDH} incorporates both anchor history $\mathcal{A}^{\prime}_{-}$ and partially-constrained referring action $\mathcal{P}^{a}$ via the trajectory-steering strategy to yield a sparse yet globally consistent anchor sequence $\mathcal{A}^{\prime}=\{a^{\prime}_j\}_{j=1}^{N_1}$, in which $\{a^{\prime}_j\}_{j=1}^{i-1}$ is anchor history $\mathcal{A}^{\prime}_{-}$, $\{a^{\prime}_j\}_{j=i}^{i+1}$ is interpolated by the \texttt{LDH} through a learnable, temporal-dependent interpolation strategy, and $\{a^{\prime}_j\}_{j=i+1}^{N_1}$ is discarded and will be updated at upcoming inference step. At inference step $i\!+\!1$, the anchor history $\mathcal{A}^{\prime}_{-}$ is updated by appending the newly generated anchor $a_{i+1}^{\prime}$ at step $i$, and the entire policy model repeats, continuously updating $\mathcal{A}$.}
    \label{fig:overview}
\end{figure*}
\section{Methodology}

% In the literature, Diffusion Policy~\cite{chi2023diffusion} imitates expert demonstrations for low-level execution, yet 
% % its ``condition-and-generate" paradigm cannot reliably respond to the external reasoning model.
% it struggles 严格到达某个由 external reasoning models 特定的位置.
While Diffusion Policy~\cite{chi2023diffusion} effectively imitates expert demonstrations for low-level execution, it struggles to precisely reach the specified target point provided by external reasoning planners.
Our goal is to develop a diffusion-based policy that can effectively incorporate external guidance provided by humans or high-level reasoning planners, generating trajectories online that accurately satisfy complex spatial constraints specified in the guidance.

\subsection{Problem Formulation}
We consider the problem of learning a diffusion-based policy \(\pi_\theta\). Given a 3D referring point \(\mathcal{P}\) (from a high-level reasoning planner such as RoboRefer~\cite{zhou2025roborefer}), the current visual observation \(\mathcal{O}\) and the robot proprioception history \(\mathcal{A}_{-}\), \(\pi_\theta\) generates a trajectory \(\mathcal{A}\) satisfying the criteria collectively termed \emph{Manipulation Steering Choreography}.
\begin{equation}
    \pi_\theta(\mathcal{P}, \mathcal{O}, \mathcal{A}_{-}) \xrightarrow{} \mathcal{A}
\end{equation}

\noindent
\textbf{Manipulation Steering Choreography.}
Guided by the high-level reasoning planner, the end-effector must glide through the 3D referring point $\mathcal{P}$ while guaranteeing successful task completion. Thus, we require every generated trajectory to satisfy:
1) \emph{Steering Fidelity}: the trajectory must establish contact within the spatial region nearby the specific 3D referring point.
2) \emph{Task Success}: the trajectory must accomplish the designated manipulation task.
3) \emph{Smoothness}: the end-effector pose must vary uniformly, yielding low-jerk transitions along the entire trajectory.

\noindent
\textbf{Referring-Aware Policy Model.}  
To effectively handle the OOD situations, we propose \mname{}, described below in two parts.
\textbf{Part 1}:
We first build a \emph{policy model with coupled diffusion heads} to learn standard task execution patterns from demonstrations: its global diffusion head (\texttt{GDH}) produces sparse action anchors encoding long-range motion intent, while the local diffusion head (\texttt{LDH}) interpolates these anchors into a fine-grained trajectory, conditioned on the current temporal position for each task.
During inference, the entire model is invoked repeatedly, enabling the robot to replan as the scene changes.
\textbf{Part 2}:
Subsequently, we embed a \emph{referring-aware design} into the aforementioned model, which centers on a temporal-position prediction module as shown in Fig.~\ref{fig:overview} and a trajectory-steering strategy. The former assigns a temporal position along the entire trajectory to the 3D referring point based on the robot current observation and robot proprioception history. And the latter injects this temporally-positioned point into our policy model, enabling the generation of a trajectory that fulfills the \emph{Manipulation Steering Choreography}.

\subsection{Policy Model with Coupled Diffusion Heads}
\label{sec:closed-loop_global_policy}
To generate a trajectory that adheres to the \emph{Manipulation Steering Choreography}, we must schedule the referring point $\mathcal{P}$ coherently across the entire manipulation trajectory, which demands a global understanding of the long-range task structure. While  previous methods~\cite{chi2023diffusion, ma2025cdp, Ze2024DP3, zhao2023learningfinegrainedbimanualmanipulation} model “observation–policy" coupling only within a sliding-window horizon, their local scope prevents them from capturing the full-horizon dependencies essential for this purpose.
% Such a scheme disrupts trajectory consistency and makes it difficult to schedule the referring points $\mathcal{P}$ along the entire manipulation trajectory.  
To overcome this limitation, we propose a policy model with coupled diffusion heads which can learn standard execution pattern from expert demonstration in a close-loop manner.

\noindent
\textbf{Coupled Diffusion Heads.}
In robotic manipulation, trajectories are prohibitively long, and their length grows with task complexity. Learning a single Diffusion Policy that captures execution pattern of an entire trajectory is therefore impractical. In this paper, we introduce a coupled diffusion heads:
A \texttt{GDH} is first deployed to generate a globally consistent yet temporally sparse action anchors
\(
\mathcal{A}^{\prime}=\{a^{\prime}_i\}_{i=1}^{N_1}
\) of length \(N_1\).
%
% In contrast to the keyposes adopted by ChainedDiffuser~\cite{xian2023chaineddiffuser}, our sparse sequence $\mathcal{A}^{\prime}$ represents a uniform temporal down-sampling of the entire trajectory and carries no sub-task semantics. The sole role of \texttt{GDH} is to distill the global motion pattern underlying the specific manipulation task, rather than to delineate its semantic stages.
%
As illustrated in Fig.~\ref{fig:overview}, \texttt{GDH} predicts $\mathcal{A}^{\prime}$ conditioned on the current observation $\mathcal{O}$ and the previously executed anchors $\mathcal{A}^{\prime}_{-}$, \ie
\begin{equation}
\mathcal{A}^{\prime} \sim f_{\,\texttt{GDH}}(\mathcal{A}^{\prime} | \mathcal{O}, \mathcal{A}^{\prime}_{-})
\label{eq:GDH}
\end{equation}
Subsequently, a \texttt{LDH} densifies these anchors, producing a fine-grained trajectory ready for direct deployment on the robot. Following Eq.~\eqref{eq:GDH}, once the neighboring anchor pair $(a_i^{\prime}, a_{i+1}^{\prime})$ for step~$i$ is available, \texttt{LDH} interpolates between them conditioned on the corresponding observation~$\mathcal{O}_i$, \ie
\begin{equation}
\mathcal{A}_i \sim f_{\,\texttt{LDH}}(\mathcal{A}_i \mid \mathcal{O}_i, a_i^{\prime}, a_{i+1}^{\prime}, i)
\label{eq:LDH}
\end{equation}
where $\mathcal{A}_i=\{a_{ij}\}_{j=1}^{N_2}$ is the fine-grained sub-trajectory of length~$N_2$ in step~$i$, and the entire manipulation trajectory is simply the concatenation $\mathcal{A} = \{\mathcal{A}_i\}_{i=1}^{N_1}$.
It is worth noting that, in Eq.~\eqref{eq:LDH}, the step index~$i$ is explicitly fed into $f_{\,\texttt{LDH}}$ so that the network can learn temporal-dependent interpolation strategies that vary with the temporal position inside $\mathcal{A}^{\prime}$.
Here, we apply our trajectory-steering strategy (Sec.~\ref{sec:trajectory-steering}) to (i) historical anchors $\mathcal{A}^{\prime}_{-}$ that guides \texttt{GDH} prediction and (ii) the neighboring anchor pair $(a_i^{\prime}, a_{i+1}^{\prime})$ that steers \texttt{LDH} interpolation, ensuring the generated trajectory strictly continues the already-executed motion while fulfilling the remaining task objectives.
Since the output of our policy model is a \emph{fixed-length} trajectory, we must select the total horizon $N = N_1 + (N_1 - 2) * N_2$ for each task once at the outset, according to its execution complexity.  
Fortunately, this mild restriction is easily accommodated in concurrent robotic manipulation pipelines, where an upper bound can be set without impairing deployment.

\noindent
\textbf{Closed-Loop Inference.}  
To remain robust to the dynamic environment, we invoke the entire policy model iteratively during inference.
As illustrated in Fig.~\ref{fig:overview}, the sparse action anchors $\mathcal{A}^{\prime}$ is updated online by continuously incorporating the latest observation $\mathcal{O}$ and the ever-growing anchor history $\mathcal{A}^{\prime}_{-}$.
Specifically, at step $i$ we generate $\mathcal{A}^{\prime}$ with \texttt{GDH} (Eq.~\eqref{eq:GDH}), extract the neighboring anchor pair $(a_i^{\prime},a_{i+1}^{\prime})$ and produce the fine-grained sub-trajectory $\mathcal{A}_i$ via \texttt{LDH} (Eq.~\eqref{eq:LDH}) for immediate execution. Once the corresponding execution process is completed, the executed anchor $a_{i+1}^{\prime}$ is appended to build the anchor history at step $i\!+\!1$, \ie
\begin{equation}
\mathcal{A}^{\prime}_{-,\,i+1}= \bigl\{\mathcal{A}^{\prime}_{-,\,i},\, a_{i+1}^{\prime}\bigr\}
\end{equation}
where $\mathcal{A}^{\prime}_{-,\,i} = \{a^{\prime}_j\}_{j=1}^{i}$ is the anchor history at step $i$. This $\mathcal{A}^{\prime}_{-,\,i+1}$ is then used to update anchors $\mathcal{A}^{\prime}$ at step $i\!+\!1$.

\subsection{Referring-Aware Design}
\label{sec:trajectory-steering}
We now detail how to embed the referring-aware design into the aforementioned policy model in order to schedule the referring point $\mathcal{P}$ along the entire trajectory.

\noindent
\textbf{Temporal-Position Prediction.}
\begin{figure}[t]
    \centering
    \setlength{\fboxrule}{0pt}
    \framebox{{\includegraphics[width=0.95\linewidth]{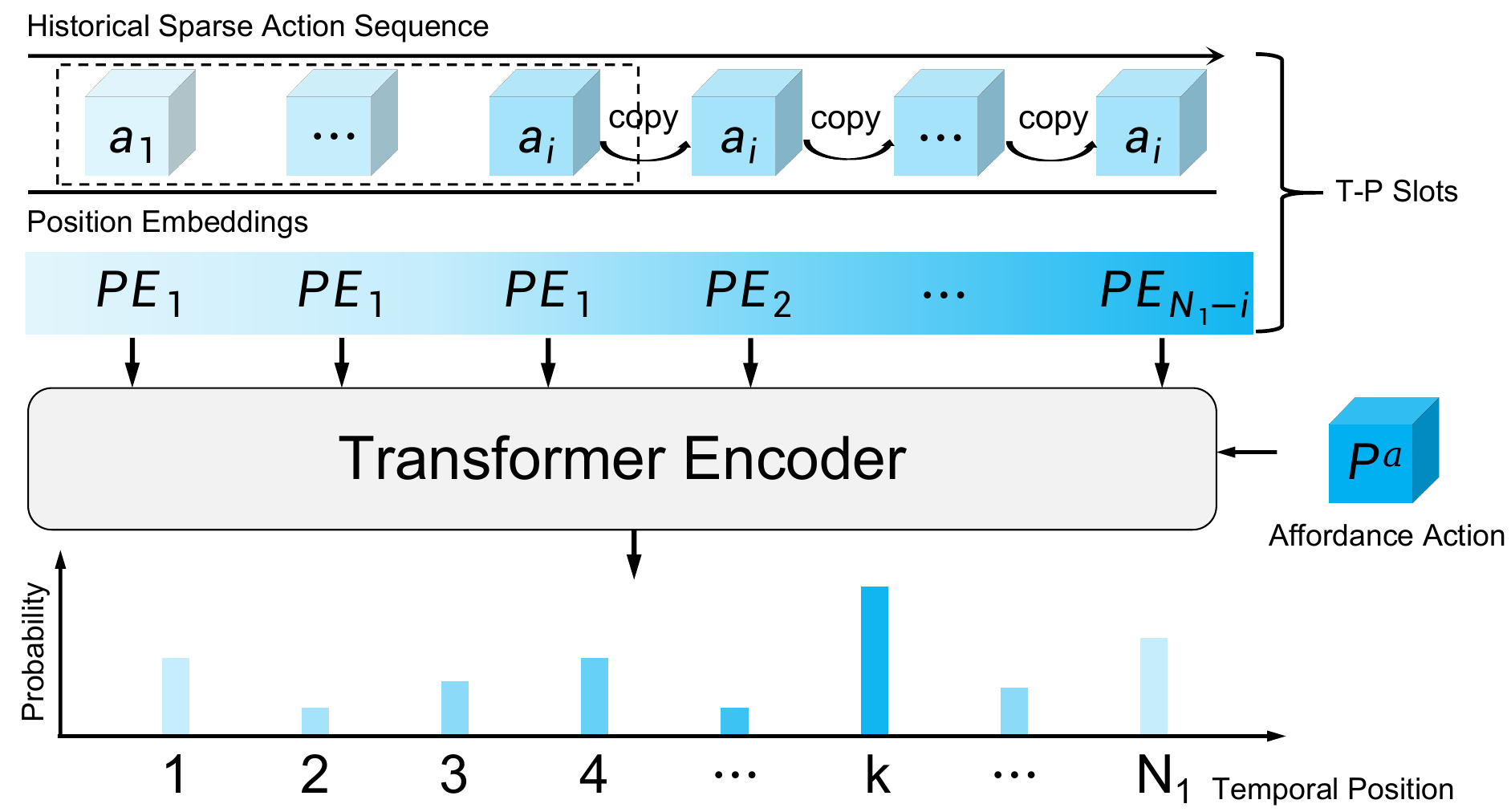}}}
    \caption{\textbf{Temporal-Position Prediction}. The slots buffer $\mathcal{S}$ covers both retained history and extrapolated future steps. Specifically, at step $i$, we load the historical action anchors $\{a^{\prime}_{j}\}_{j=1}^{i}$ to initialize history part, and the future part is padded by copies of the last action anchor $a^{\prime}_{i}$. Then, the whole slots buffer is augmented with a monotonically-increasing temporal-position embeddings.}
    \label{fig:temporal_position_prediction}
\end{figure}
As shown in Fig.~\ref{fig:temporal_position_prediction}, we formulate the temporal localization of the referring point $\mathcal{P}$ as an $N_{1}$-way classification problem.
Concretely, we construct a \emph{fixed-length} timeline by creating a buffer of $N_{1}$ temporal-position slots (\texttt{T-P Slots}) $\mathcal{S}$, covering both retained history and extrapolated future steps.
In step $i$, we build $\mathcal{S}$ as follows: The first $(1,\dots,i)$ slots are loaded with the historical action anchors $\{a^{\prime}_{j}\}_{j=1}^{i}$, and the remaining $(i\!+\!1,\dots,N_{1})$ slots are padded with copies of the last action anchor $a^{\prime}_{i}$, yielding a slot sequence \(\,[a^{\prime}_{1},\,a^{\prime}_{2},\,\dots,\,a^{\prime}_{i},\,\dots,\,a^{\prime}_{i}\,]\). This slot sequence is then augmented with the monotonically-increasing temporal-position embeddings. Specifically, the historical slots $(1,\dots,i)$ share the same temporal-position embedding $P\!E_{1}$, while the padded slots $(i\!+\!1,\dots,N_{1})$ receive distinct embeddings $P\!E_{2}\!<\!P\!E_{3}\!<\!\dots\!<\!P\!E_{N_{1}-i}$. These embeddings encode the temporal distance from the action in each slot to the referring point $\mathcal{P}$: $P\!E_{1}$ marks the nearest moment, and larger embeddings indicate progressively more distant future times.
A transformer-based encoder processes the augmented $\mathcal{S}$ together with $\mathcal{P}$ and outputs a probability vector $\mathbf{p} = \{p_k\}_{k=1}^{N_1}$ over the entire slots buffer.  
The predicted temporal position is taken as the slot index with the highest probability, \ie
\begin{equation}
k = \arg\max_{0 < k \leq N_1} p_{k}
\label{eq:TP}
\end{equation}
% \begin{figure}[t]
%     \centering
%     \setlength{\fboxrule}{0pt}
%     \framebox{{\includegraphics[width=0.8\linewidth]{fig/Policy-Editing.png}}}
%     \caption{\textbf{Policy Editing.}}
%     \label{fig:policy-editing}
% \end{figure}

\noindent
\textbf{Trajectory-Steering Strategy.} Before detailing its mechanics, it is worth noting that all states and actions are represented in \emph{end-effector Cartesian space}, rather than in \emph{joint space}. Formally, the actions considered in this paper are decomposed into:
1) an end-effector pose component \(a_{\texttt{ee}}=(a_{\texttt{trans}},a_{\texttt{rot}})\), where \(a_{\texttt{trans}}\in\mathbb{R}^{3}\) and \(a_{\texttt{rot}}\in\mathbb{R}^{4}\) denote the position and rotation respectively.
2) a binary gripper component \(a_{\texttt{gripper}}\in\{0,1\}\) that opens or closes the parallel jaw.
This representation allows us to convert the constraint on the referring point $\mathcal{P}$ into a \emph{partial} constraint on the referring action $\mathcal{P}^{a}$: we simply force its translation component \(a_{\texttt{trans}}\) to coincide with the referring point $\mathcal{P}$, while leaving the rotation component \(a_{\texttt{rot}}\) free to be optimized by the policy model. We then implement our \emph{trajectory-steering} strategy through a masked-denoising process~\cite{tseng2023edge, kim2023flame}, \ie
\begin{equation}
z_{t} = \mathcal{M} \odot A_\texttt{known} + (1 - \mathcal{M}) \odot z_{t}
\label{eq:trajectory-steering}
\end{equation}
where $z_{t}$ denotes the intermediate noisy action vector at diffusion timestep $t$, $\odot$ represents the Hadamard (element-wise) product, $A_\texttt{known}$ is a known action vector used to steer the denoising, and $\mathcal{M}$ is a binary mask indicating the indices to replace within the full noisy trajectory.
As stated in Sec.~\ref{sec:closed-loop_global_policy}, this strategy is applied in two stages: (i) sparse-anchor $\mathcal{A}^{\prime}$ generation in \texttt{GDH},
\begin{equation}
z_{t} = \mathcal{M}_\texttt{GDH} \odot \mathcal{A}^{\prime}_{-} + (1 - \mathcal{M}_\texttt{GDH}) \odot z_{t}
\label{eq:trajectory-steering_GDH}
\end{equation}
and (ii) anchor interpolation in \texttt{LDH},
\begin{equation}
z_{t} = \mathcal{M}_\texttt{LDH} \odot (a_i^{\prime}, a_{i+1}^{\prime}) + (1 - \mathcal{M}_\texttt{LDH}) \odot z_{t}
\label{eq:trajectory-steering_LDH}
\end{equation}
used to generate the fine-grained sub-trajectory $\mathcal{A}_i$. Furthermore, in our referring-aware design, we incorporate the referring action $\mathcal{P}^{a}$ to Eq.~\eqref{eq:trajectory-steering_GDH} as follows,
\begin{equation}
z_{t} = \mathcal{M}^{\prime}_\texttt{GDH} \odot \{\mathcal{A}^{\prime}_{-}, \mathcal{P}^{a}\} + (1 - \mathcal{M}^{\prime}_\texttt{GDH}) \odot z_{t}
\label{eq:trajectory-steering_GDH_prime}
\end{equation}
thereby guiding the generation toward trajectories that satisfy the \emph{Manipulation Steering Choreography}. Here, $\mathcal{M}_\texttt{GDH}$, $\mathcal{M}_\texttt{GDH}$ and $\mathcal{M}^{\prime}_\texttt{GDH}$ denote the binary masks corresponding to $\mathcal{A}^{\prime}_{-}$, $(a_i^{\prime}, a_{i+1}^{\prime})$, $\{\mathcal{A}^{\prime}_{-}, \mathcal{P}^{a}\}$, respectively. Detailed definitions of these masks are provided in Appendix.~\ref{sec:binary_mask}.

% To allow our policy model to attend to both the anchor history $\mathcal{A}^{\prime}_{-}$ and the referring action $\mathcal{P}^{a}$, we introduce the referring action $\mathcal{P}^{a}$ into Eq.~\eqref{eq:GDH}, \ie
% \begin{equation}
% \mathcal{A}^{\prime} \sim f_{\,\texttt{GDH}}(\mathcal{A}^{\prime} | \mathcal{O}, \mathcal{A}^{\prime}_{-}, \mathcal{P}^{a})
% \end{equation}

\subsection{Training  Strategy}
\noindent
\textbf{Data Recipe.}  
To enlarge the effective range of $\mathcal{P}$ and promote generalization, we perform on-the-fly augmentation: we randomly sample an action from the expert demonstrations, perturb it with noise drawn from a broad distribution, and obtain a synthetic referring action $\mathcal{P}^{a}$.  
A seventh-order polynomial spline then smoothly blends this synthetic action with its temporal neighborhood, yielding a jerk-bounded trajectory.  
These augmented trajectories are fed to both \texttt{GDH} and \texttt{LDH}, enabling the system to handle a richer spectrum of referring points $\mathcal{P}$ at inference time.

\noindent
\textbf{Temporal-Position Prediction Loss.}  
The transformer-based encoder used for temporal-position prediction is trained with the following categorical cross-entropy loss
\begin{equation}
\mathcal{L}_{\texttt{CCE}}=-\sum_{i=1}^{N_{1}}y_{i}\log p_{i}
\end{equation}
where $y_{i}\in\{0,1\}$ is the ground-truth that indicates whether slot $i$ corresponds to the referring action $\mathcal{P}^{a}$.  

\noindent
\textbf{Coupled Diffusion Heads Loss.}  
Besides, the demonstration trajectory~$\hat{\mathcal{A}}$ is first down-sampled to yield the sparse anchors label~$\hat{\mathcal{A}}^{\prime}$.  
Using the down-sampling indices, $\hat{\mathcal{A}}$ is then split into $N_1$ contiguous segments~$\{\hat{\mathcal{A}}_{i}\}_{i=1}^{N_1}$, which serve as labels for \texttt{LDH} supervision. The loss
\begin{equation}
\mathcal{L}_{\texttt{MSE}} = 
\mathbb{E}\!\bigl[\|\mathcal{A}^{\prime}-\hat{\mathcal{A}}^{\prime}\|_{2}^{2}\bigr]
\;+\;
\gamma\,\mathbb{E}\!\bigl[\|\mathcal{A}_{i}-\hat{\mathcal{A}}_{i}\|_{2}^{2}\bigr]
\end{equation}
jointly supervises \texttt{GDH} and \texttt{LDH}, where the scalar $\gamma$ balances their relative importance.  
During training, \texttt{LDH} is updated by uniformly sampling one segment index $i$ per iteration.

\noindent
\textbf{Total Loss.} 
The overall training objective is
\begin{equation}
\mathcal{L}=\mathcal{L}_{\texttt{CCE}}+\alpha\,\mathcal{L}_{\texttt{MSE}}
\end{equation}
with the scalar hyper-parameter $\ \alpha$ balancing the two terms.

\begin{table*}[t!]
    \centering
    \renewcommand{\arraystretch}{1.5}
    \caption{\textbf{Quantitative results on modified simulated benchmark}, highlighting the effectiveness of \mname{} in referring-aware manipulation.
    }
    \label{tab:controllable_quality}
    \resizebox{0.96\textwidth}{!}{
    \begin{tabular}{c c c c c c c c c c c c c}
        \toprule[1pt]
        \multirow{2}{*}{Method} & \multicolumn{3}{c}{Pick Meat-via} & \multicolumn{3}{c}{Lift Barrier-via} & \multicolumn{3}{c}{Place Food-via} & \multicolumn{3}{c}{Camera Alignment-via} \\ \cmidrule(lr){2-4} \cmidrule(lr){5-7} \cmidrule(lr){8-10} \cmidrule(lr){11-13} 
        & \texttt{RePR(}$\uparrow$\texttt{)} & \texttt{SuR(}$\uparrow$\texttt{)} & \texttt{SmS(}$\uparrow$\texttt{)} & \texttt{RePR(}$\uparrow$\texttt{)} & \texttt{SuR(}$\uparrow$\texttt{)} & \texttt{SmS(}$\uparrow$\texttt{)} & \texttt{RePR(}$\uparrow$\texttt{)} & \texttt{SuR(}$\uparrow$\texttt{)} & \texttt{SmS(}$\uparrow$\texttt{)} & \texttt{RePR(}$\uparrow$\texttt{)} & \texttt{SuR(}$\uparrow$\texttt{)} & \texttt{SmS(}$\uparrow$\texttt{)} \\ \hline 
        ACT & 2\% & 1\% & 0.9890 & 1\% & 1\% & 0.9904 & 0\% & 0\% & - & 0\% & 0\% & - \\
        DP3 & 80\% & 1\% & 0.9899 & 99\% & 25\% & \textbf{0.9945} & 1\% & 1\% & \textbf{0.9883} & 0\% & 0\% & -  \\
        CDP & 14\% & 14\% & \textbf{0.9924} & 99\% & 99\% & 0.9933 & 47\% & 33\% & 0.9878 & 0\% & 0\% & -  \\

        OCTO & 18\% & 9\% & 0.9606  & 32\% & 32\% & 0.9702 & 1\% & 1\% & 0.9597 & 0\% & 0\% & -  \\ 

        MPD & 20\% & 3\% & 0.9903  & 39\% & 39\% & 0.9949 & 3\% & 3\% & 0.9887 & 1\% & 1\% & \textbf{0.9861}  \\ \hline 
        
        \mname{}~(Linear) & \textbf{100\%} & 80\% & 0.9875  & \textbf{100\%} & 63\% & 0.9867 & \textbf{100\%} & 21\% & 0.9834 & \textbf{100\%} & 87\% & 0.9760 \\ 
        
        \mname{}~(Cubic Spline) & \textbf{100\%} & 85\% & 0.9907  & \textbf{100\%} & 86\% & 0.9927 & \textbf{100\%} & 17\% & 0.9821 & \textbf{100\%} & 85\% & 0.9810 \\  
        
        \mname{}~(Minimum Snap) & \textbf{100\%} & 18\% & 0.9855  & \textbf{100\%} & 80\% & 0.9914 & \textbf{100\%} & 23\% & 0.9828 & \textbf{100\%} & 76\% & 0.9799 \\ 
        
        \rowcolor{lightblue} 
        \mname{} & \textbf{100\%} & \textbf{91\%} & 0.9882  & \textbf{100}\% & \textbf{100\%} & 0.9899 & \textbf{100}\% & \textbf{50\%} & 0.9812 & \textbf{100\%} & \textbf{92\%} & 0.9804  \\ 
        \bottomrule[1pt]
    \end{tabular}
    }
\end{table*}
\section{Experiments}
This section evaluates the ability of our \mname{} to respond to the referring point $\mathcal{P}$ provided by the human or high-level planner. Through a series of experiments, we investigate the following questions:
\textbf{Q1.} Does our \mname{} outperform other visuomotor policies in referring-aware manipulation?
\textbf{Q2.} How does \mname{} compare to prior representative conditioning methods in accurately adhering to the provided referring point?
\textbf{Q3.} Is our \mname{} robust to referring points deviating from the expert trajectory distribution? 
\textbf{Q4.} Does the proposed Couple Diffusion Heads architecture (which captures long-horizon task execution motion) improve task success, independent of referring awareness?
\textbf{Q5.} For generating dense action trajectories between anchors, does the learnable \texttt{LDP} outperform traditional interpolation and constrained optimization methods in robotic manipulation domain?
\textbf{Q6.} Which design decisions in \mname{} matter most for building robust referring-aware policies? 
\textbf{Q7.} Can \mname{} be successfully deployed in real-world settings?
% , and how does it handle dynamic, unstructured environments?

\subsection{Referring-Awareness Evaluation}
\label{exp:trajectory-steering}
\begin{figure*}[t]
    \centering
    \setlength{\fboxrule}{0pt}
    \framebox{{\includegraphics[width=0.9\linewidth]{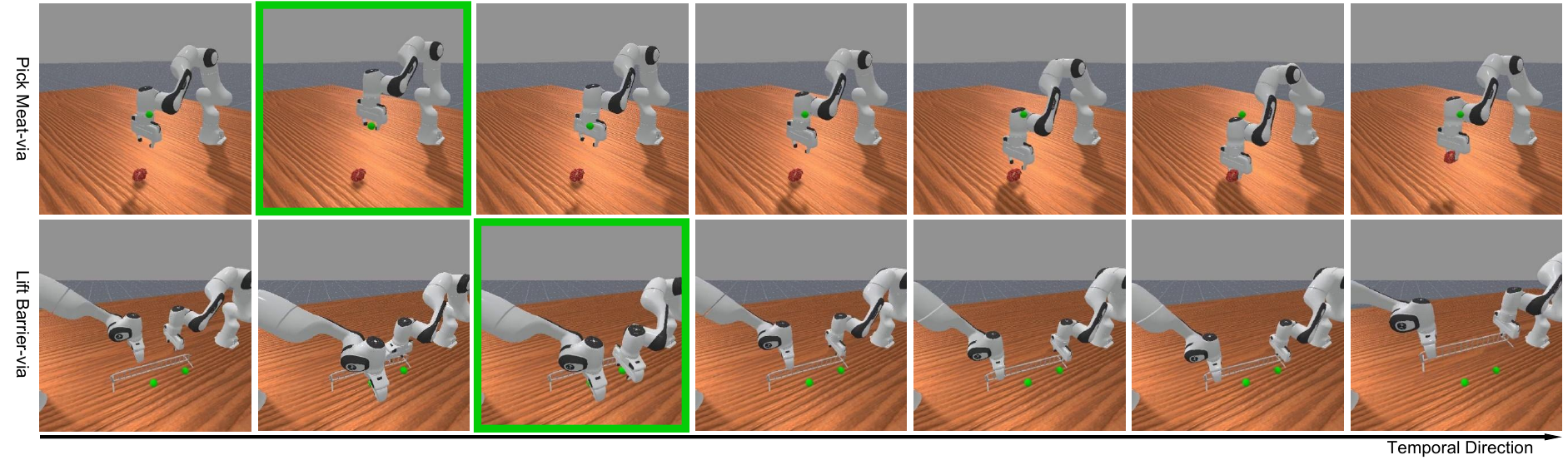}}}
    \caption{\textbf{Visualization of the trajectories generated by \mname{}} on \emph{Pick-Meat-via} and \emph{Lift-Barrier-via}. Here, we use green bounding boxes to mark the frames in which the end-effector passes through the designated via-point (green ball).}
    \label{fig:quality_results}
\end{figure*}

\noindent
\textbf{Evaluation Metrics.}
According to the \emph{Manipulation Steering Choreography}, we derive the following three metrics, each averaged over $M$ independent roll-outs.
\begin{itemize}
\item \textit{Region Penetration Rate} (\texttt{RePR})
measures the fraction of trajectories in which the robot's end-effector passes through the referring point $\mathcal{P}$. For trajectory~$i$, let  
\begin{equation}
d_i = \min_{0 < t \le N} \| \mathbf{p}_i(t) - \mathcal{P} \|_2
\end{equation}
denote the minimum Euclidean distance between the end-effector position $\mathbf{p}_i(t)$ and $\mathcal{P}$ over the entire trajectory ($0 < t \le N$). We count a penetration if $d_i\le\epsilon$ and define
\begin{equation}
\texttt{RePR}=\frac{1}{M}\sum_{i=1}^{M} \mathbb{I}\bigl[d_i \le \epsilon \bigr]
\end{equation} 
The threshold~$\epsilon$ specifies the penetration tolerance, and in our experiments it is fixed at $\SI{0.05}{\meter}$. 
% $\mathbb{I}(\cdot)$ denotes the indicator function, which equals $1$ if the condition inside is true and $0$ otherwise.

\item \textit{Success Rate} (\texttt{SuR}).  
In contrast to \cite{Ze2024DP3,ma2025cdp,su2025dense}, a roll-out is considered successful in this experiment only if the robot completes the assigned task and its end-effector passes through the designated referring point $\mathcal{P}$ during execution, \ie
\begin{equation}
\texttt{SuR}=\frac{1}{M}\sum_{i=1}^{M} \mathbb{I}\bigl[d_i\le\epsilon\land S_i=1\bigr]
\end{equation}
where $S_i\!\in\!\{0,1\}$ is the task-completion label.

\item \textit{Smoothness Score} (\texttt{SmS}).  
For trajectory $i$, we compute  
\begin{equation}
J_i = \frac{1}{N-1}\sum_{t=1}^{N-1} \,\bigl\|\mathbf{p}_{i,t+1}-\mathbf{p}_{i,t}\bigr\|_2
\end{equation}
To map the unbounded $J_i$ to $[0,\,1]$, we use  
\begin{equation}
s_i=\exp(-J_i/\lambda)
\end{equation}
with temperature $\lambda\!>\!0$, thus smooth trajectories yield $s_i\!\approx\!1$ and jittery ones $s_i\!\approx\!0$. Subsequently, we define
\begin{equation}
\texttt{SmS}=\frac{1}{M^{\prime}}\sum_{i=1}^{M^{\prime}}s_i
\end{equation}
where $M^{\prime}$ denotes the number of roll-outs that satisfy the success criterion used in the definition of $\texttt{SuR}$.
\end{itemize}

\noindent
\textbf{Baselines.}  
We select four representative conditioning policies—ACT~\cite{zhao2023learningfinegrainedbimanualmanipulation}, DP3~\cite{Ze2024DP3}, CDP~\cite{ma2025cdp} and Octo~\cite{octo_2023}, MPD~\cite{carvalho2025motion}—as baselines.
For the first three baseline methods, we concatenate the referring point $\mathcal{P}$ directly with the visual and proprioception observations as an additional condition. For the forth approach, we embed $\mathcal{P}$ as a natural language instruction to condition the generative model. As for the final method, we formulate a guidance cost based on the referring point $\mathcal{P}$, which steers the denoising process via classifier-guided sampling.
% All baselines are designed and conditioned with the goal of generating trajectories that satisfy the \emph{Manipulation Steering Choreography}. 
% All models are trained on the same expert demonstrations for an identical number of epochs to ensure experimental parity.

\noindent
\textbf{Benchmarks.} 
We augment four representative tasks from RoboFactory~\cite{qin2025robofactory}—\emph{Pick Meat}, \emph{Lift Barrier}, \emph{Place Food}, and \emph{Camera Alignment}—by introducing a via-point that the robot's end-effector must traverse en route to successful completion. This yields the modified benchmark suite: \emph{Pick Meat-via}, \emph{Lift Barrier-via}, \emph{Place Food-via}, and \emph{Camera Alignment-via}. The via-point generation strategy is detailed in Appendix~\ref{sec:benchmark_modification}.

\noindent
\textbf{Quantitative and Qualitative Results (Q1).}  
Tab.~\ref{tab:controllable_quality} shows that our \mname{} yields the highest proportion of trajectories that satisfy \emph{Manipulation Steering Choreography}. Thanks to the trajectory-steering strategy, our \mname{} guarantees that 100\% of roll-outs pass through the designated referring point $\mathcal{P}$ (\cf\texttt{RePR}); and its success rate \texttt{SuR} is mainly influenced by the capability of policy model (\cf Tab.~\ref{tab:hierarchical-framework}).
In contrast, the baselines overwhelmingly ignores $\mathcal{P}$ and proceeds directly to the final goal, exposing its weakness in referring-aware manipulation.  
Fig.~\ref{fig:quality_results} overlays representative trajectories produced by our \mname{} across the aforementioned tasks, which simultaneously accomplishes the task while smoothly traversing $\mathcal{P}$, demonstrating markedly superior referring-awareness.

\subsection{Ablation Study}
\noindent
\textbf{Fidelity to Referring Points (Q2).} 
As shown in Tab.~\ref{tab:controllable_quality}, while baseline methods largely ignore the provided guidance, our \mname{} strictly follows the referring point $\mathcal{P}$, thereby maintaining high precision in reaching them (\cf \texttt{RePR}).
To further verify that our \mname{}’s behavior is causally governed by the provided referring points, we deliberately introduce infeasible referring points—guidance signals that contradict successful task completion. The design of these infeasible points is detailed in Appendix~\ref{sec:design_of_infeasible_referring_points}. And its corresponding quantitative results as listed in Tab.~\ref{tab:infeasible_referring_points}.
Specifically, in the first two tasks, our \mname{} physically pushes aside the camera or pot to reach the point, achieving \texttt{RePR} = 100\%. In contrast, for the latter two tasks, the robot fails to reach the points due to physical constraints (occlusion or workspace limits), leading to \texttt{RePR} = 0\%. 
% It demonstrates that our referring mechanism can provide genuine, actionable control over the generated trajectory, reliably steering it—even toward failure—when the guidance is infeasible.

\begin{table}[t]
    \centering
    \renewcommand{\arraystretch}{1.5}
    \caption{\textbf{Ablation study on OOD-yet-feasible referring points.} 0.1, 0.2, 0.3, and 0.4 indicate the degree of deviation from the center of the expert distribution. Details are provided in Appendix~\ref{sec:design_of_out-of-distribution_referring_points}.
    }
    \label{tab:ood_referring_points}
    \resizebox{0.48\textwidth}{!}{
    \begin{tabular}{c c c c c c c c c}
        \toprule[1pt]
        \multirow{2}{*}{Method} & \multicolumn{2}{c}{0.1} & \multicolumn{2}{c}{0.2} & \multicolumn{2}{c}{0.3} & \multicolumn{2}{c}{0.4} \\ \cmidrule(lr){2-3} \cmidrule(lr){4-5} \cmidrule(lr){6-7} \cmidrule(lr){8-9} 
        & \texttt{RePR(}$\uparrow$\texttt{)} & \texttt{SuR(}$\uparrow$\texttt{)} & \texttt{RePR(}$\uparrow$\texttt{)} & \texttt{SuR(}$\uparrow$\texttt{)} &  \texttt{RePR(}$\uparrow$\texttt{)} & \texttt{SuR(}$\uparrow$\texttt{)} &  \texttt{RePR(}$\uparrow$\texttt{)} & \texttt{SuR(}$\uparrow$\texttt{)} \\
        \hline 
        \rowcolor{lightblue}
        \mname{} & 100\% & 93\% & 100\% & 92\% & 100\% & 89\% & 100\% & 87\% \\ 
        \bottomrule[1pt]
    \end{tabular}
    }
    % \vspace{-\baselineskip}
\end{table}
\begin{table*}[t]
    \centering
    \renewcommand{\arraystretch}{1.5}
    \caption{\textbf{Quantitative results across simulated benchmarks}, highlighting the effectiveness of our Coupled Diffusion Heads architecture.
    }
    \label{tab:hierarchical-framework}
    \resizebox{\textwidth}{!}{
    \begin{tabular}{c c c c c c c c c c c c c c}
        \toprule[1pt]
        \multirow{2}{*}{Method} & \multicolumn{2}{c}{Adroit} & \multicolumn{3}{c}{DexArt} & \multicolumn{4}{c}{MetaWorld} & \multicolumn{4}{c}{RoboFactory} \\ \cmidrule(lr){2-3} \cmidrule(lr){4-6} \cmidrule(lr){7-10} \cmidrule(lr){11-14}
        & Pen & Door & Laptop & Toilet & Bucket & Reach & Soccer & Sweep Into & Shelf Place & Pick Meat & Lift Barrier & Place Food & Camera Alignment \\ \hline 
        ACT & 47\% & 66\% & 35\% & 8\% & 6\% & 21\% & 28\% & 23\%  & 43\% & 91\% & 40\% & 13\% & 21\%\\  
        DP3 & 53\% & 69\% & 81\% & 65\% & 32\% & 28\% & 30\% & 15\% & 37\% & 81\% & 90\% & 30\% & 87\% \\ 
        CDP & 68\% & 74\% & 84\% & 68\% & 32\% & 22\% & 23\% & 24\% & 35\% & 84\% & 93\% & 51\% & 90\% \\ 
        \rowcolor{lightblue} 
        \mname{} & \textbf{73\%} & \textbf{79\%} & \textbf{87\%} & \textbf{71\%} & \textbf{61\%} & \textbf{36\%} & \textbf{33\%} & \textbf{27\%} & \textbf{47\%} & \textbf{94\%} & \textbf{99\%} & \textbf{57\%} & \textbf{94\%} \\ 
        \bottomrule[1pt]
    \end{tabular}
    }
\end{table*}
\begin{table*}[t!]
    \centering
    \renewcommand{\arraystretch}{1.5}
    \caption{\textbf{Quantitative results on real-world tasks}, highlighting the robustness of our \mname{} in real-world settings.}
    \label{tab:real-performance}
    \resizebox{0.8\textwidth}{!}{
    \begin{tabular}{c c c c c c c c c c c}
        \toprule[1pt]
        \multirow{3}{*}{Method} & 
        \multicolumn{6}{c}{Single-Agent} & \multicolumn{4}{c}{Dual-Agent} \\
        \cmidrule(lr){2-7} \cmidrule(lr){8-11} & 
        \multicolumn{2}{c}{Collecting Objects-via} & 
        \multicolumn{2}{c}{Push T-via} & 
        \multicolumn{2}{c}{Stacking Playing Card-via} & 
        \multicolumn{2}{c}{Grabbing Rod-via} &
        \multicolumn{2}{c}{Handing Eraser Over-via} \\ 
        \cmidrule(lr){2-3} \cmidrule(lr){4-5} \cmidrule(lr){6-7}
        \cmidrule(lr){8-9} \cmidrule(lr){10-11}
        & \texttt{PeRP}~$\uparrow$ & \texttt{SuR}~$\uparrow$ & 
        \texttt{PeRP}~$\uparrow$ & \texttt{SuR}~$\uparrow$ & 
        \texttt{PeRP}~$\uparrow$ & \texttt{SuR}~$\uparrow$ & 
        \texttt{PeRP}~$\uparrow$ & \texttt{SuR}~$\uparrow$ & 
        \texttt{PeRP}~$\uparrow$ & \texttt{SuR}~$\uparrow$ \\ 
        \midrule
        ACT & 3 / 30 & 1 / 30 & 5 / 30 & 3 / 30 & 3 / 30 & 0 / 30 & 2 / 30 & 0 / 30 & 3 / 30 & 1 / 30 \\
        DP & 6 / 30 & 5 / 30 & 12 / 30 & 9 / 30 & 9 / 30 & 3 / 30 & 3 / 30 & 1 / 30 & 6 / 30 & 2 / 30 \\
        \rowcolor{lightblue}
        \mname{} & \textbf{30 / 30} & \textbf{20 / 30} & \textbf{30 / 30} & \textbf{21 / 30} & \textbf{30 / 30} & \textbf{15 / 30} & \textbf{30 / 30} & \textbf{18 / 30} & \textbf{30 / 30} & \textbf{12 / 30} \\
        \bottomrule[1pt]
    \end{tabular}
    }
\end{table*}
\noindent
\textbf{Generalization to Out-of-Distribution Referring Points (Q3).} 
We conduct an ablation study to evaluate the robustness of our \mname{} to OOD referring points. In contrast to the deliberately infeasible points in Tab.~\ref{tab:infeasible_referring_points}, all referring points in this ablation are feasible for task completion despite being OOD, allowing us to isolate the impact of distribution shift from inherent infeasibility. The design of these OOD-yet-feasible points is detailed in Appendix~\ref{sec:design_of_out-of-distribution_referring_points}. As shown in Tab.~\ref{tab:ood_referring_points}, while the performance of our \mname{} gracefully degrades as the referring points deviate further from the expert trajectory distribution, it still maintains a high success rate, demonstrating strong generalization capability under significant distributional shift.

\noindent
\textbf{Effectiveness of our Coupled Diffusion Heads Architecture (Q4).} Tab.~\ref{tab:hierarchical-framework} quantitatively shows that our Coupled Diffusion Heads architecture consistently outperforms baseline methods across a diverse set of grasping tasks. These results highlight the decisive importance of the global execution pattern for successful task completion: (1) by modeling long-range dependencies, our policy model ensures consistency along the entire generated trajectory; (2) this strategy endows the model with a macroscopic understanding of task-specific execution motion, thereby avoiding failure cases caused by falling into locally ambiguous robot states.

\noindent
\textbf{Effectiveness of Our Learnable Local Diffusion Head (Q5).} We ablate our learnable \texttt{LDH} against interpolation and optimization baselines after \texttt{GDH} anchor generation (\cf Tab.~\ref{tab:controllable_quality}). The results highlight that the fixed or optimization-based interpolation strategies cannot adapt to the non-uniform densification needed during the entire robotic manipulation trajectory (\eg coarse early motion vs. fine-grained later adjustments).
% The results highlight two shortcomings of these methods: (1) Fixed interpolation cannot adapt to the non-uniform densification needed during the entire robotic manipulation trajectory (\eg coarse early motion vs. fine-grained later adjustments); (2) Trajectory optimization often sacrifices strict passage through anchor points—and thus through the embedded referring points—for broader optimality, losing referring-awareness.
%
Our \texttt{LDH} overcomes both limitations. Not only is its densification strategy conditioned on the anchor’s temporal position $i$ (Eq.~\ref{eq:LDH}), enabling adaptive, non-uniform refinement across the manipulation sequence, but its implementation is also fundamentally rooted in a trajectory-steering mechanism. This ensures that the generated trajectory strictly passes through every anchor point, thereby preserving explicit referring-awareness.

\noindent
\textbf{Hyperparameters (Q6).} We ablate all hyperparameters in Appendix~\ref{sec:ablation_study_on_hyperparameters}. Key findings (Fig.~\ref{fig:ablation_total}) are: 
(1) Use moderate trajectory lengths $N$ adapted to task complexity.
(2) A balanced allocation ratio between $N_{1}$ and $N_{2}$—ranging from $1\!:\!2$ to $2\!:\!1$—generally yields the most robust results. 
(3) When referring points change abruptly (e.g., due to external disturbances or re‑planning), the system should re‑initialize by resetting our \mname{} with the robot’s current state as the new start point, re‑estimate the temporal position of the referring point, and re‑run our coupled diffusion heads to generate a consistent trajectory.

\subsection{Real-World Experiments}
\noindent
\textbf{Settings.} The real-world platform, task specifications, and demonstration data used in our real-world experiments are described in detail in Appendix~\ref{sec:real-world_experiments_details}.

\noindent
\textbf{Quantitative Results (Q7).}
For each task we constructed a fixed set of 30 diverse real-world trials, and every model was evaluated on the same 30-trial split, ensuring identical test conditions for all comparisons.  
Tab.~\ref{tab:real-performance} yield that \mname{} outperforms all baselines, demonstrating the effectiveness of our approach. Here, success rate \texttt{SuR} jointly considers penetration success and final task completion.  

\section{Conclusion}
In this paper, we introduce referring-aware visuomotor policy, a novel close-loop scheme to effectively respond to the external referring information provided by humans or high-level reasoning planners. It robustly handles out-of-distribution perturbations in dynamic environments, only trained by using expert demonstrations under the imitation learning framework without any fine-tuning in the post-processing.
%
%Concretely, we first utilize a transformer-based encoder to schedule the 3D referring points along the entire trajectory. And then, these temporal-positioned referring points are injected into our policy model with coupled diffusion head through the trajectory-steering strategy, generating the final trajectory which can accomplish the specified task which pass through the referring points. 
%
To realize this, a carefully designed policy model with coupled diffusion heads is leveraged to generate the detailed action trajectory progressively.
Extensive simulated and real-world experiments show that our method outperforms baseline methods in referring-aware manipulation.
In future work, \mname{} can be coupled with VLMs and world models to tackle more complex and flexible tasks, thanks to their awareness of the reasoning signals. 

\noindent
\textbf{Limitation.}  
The proposed ReV supports the use of multiple referring points.
However, all experiments in this paper only focus on a single referring point due to the naive architecture design of the temporal position prediction module. 
Additionally, all experiments presented in this paper focus on evaluating the model's ability to faithfully respond to the referring point, without addressing how the referring point is generated by the underlying foundational model.
In future work, we plan to investigate the aforementioned issues further to develop more robust and generalizable solutions.

% \section*{Acknowledgements}

\section*{Impact Statement}

This work aims to advance the field of embodied intelligence, with a core objective of enhancing agents' decision-making and interactive capabilities in physical environments. Given that our research involves agents interacting with the real world, we acknowledge the associated safety concerns, accountability issues, and potential risks of misuse. The experiments conducted in this study are performed in controlled settings; however, we emphasize that any future real-world deployment of such technologies must incorporate rigorous safety testing frameworks, fault-tolerant redundancy mechanisms, and human oversight to prevent unintended harm or property damage.

\nocite{langley00}

\bibliography{example_paper}
\bibliographystyle{icml2026}

%%%%%%%%%%%%%%%%%%%%%%%%%%%%%%%%%%%%%%%%%%%%%%%%%%%%%%%%%%%%%%%%%%%%%%%%%%%%%%%
%%%%%%%%%%%%%%%%%%%%%%%%%%%%%%%%%%%%%%%%%%%%%%%%%%%%%%%%%%%%%%%%%%%%%%%%%%%%%%%
% APPENDIX
%%%%%%%%%%%%%%%%%%%%%%%%%%%%%%%%%%%%%%%%%%%%%%%%%%%%%%%%%%%%%%%%%%%%%%%%%%%%%%%
%%%%%%%%%%%%%%%%%%%%%%%%%%%%%%%%%%%%%%%%%%%%%%%%%%%%%%%%%%%%%%%%%%%%%%%%%%%%%%%
\newpage
\appendix
\onecolumn
\section{Implementation Details of Our \mname{}}
This section details the experimental setup for evaluating our \mname{} across simulation and real-world environments. A comprehensive description of the baseline methods and their configurations is provided in the following sections.
\subsection{Demonstration For Training}
\begin{figure}[ht!]
    \centering
    \setlength{\fboxrule}{0pt}
    \framebox{{\includegraphics[width=0.65\linewidth]{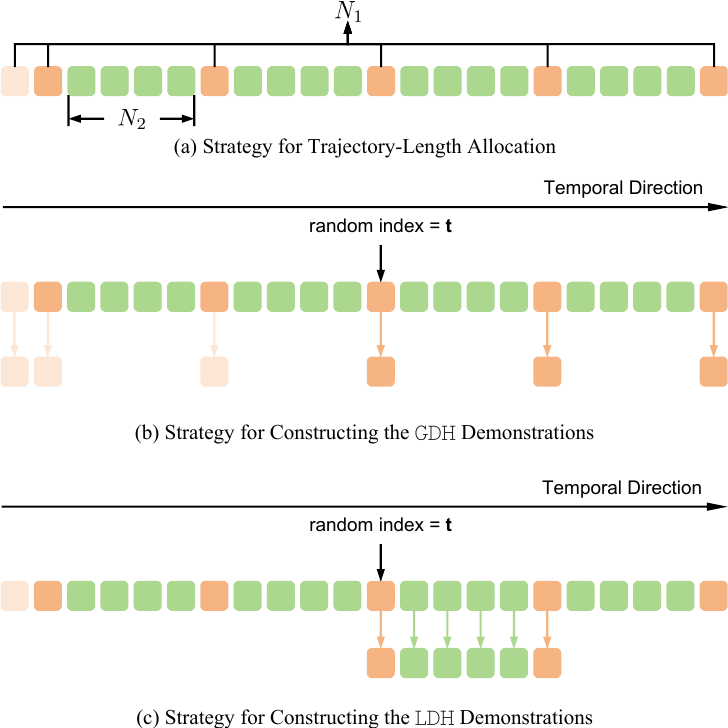}}}
   \caption{\textbf{Demonstration Construction Strategy for} \texttt{GDH} \textbf{and} \texttt{LDH}. The orange blocks represent the sparse action anchors. Among them, the lighter orange \raisebox{-0.7ex}{\includegraphics[height=0.32cm]{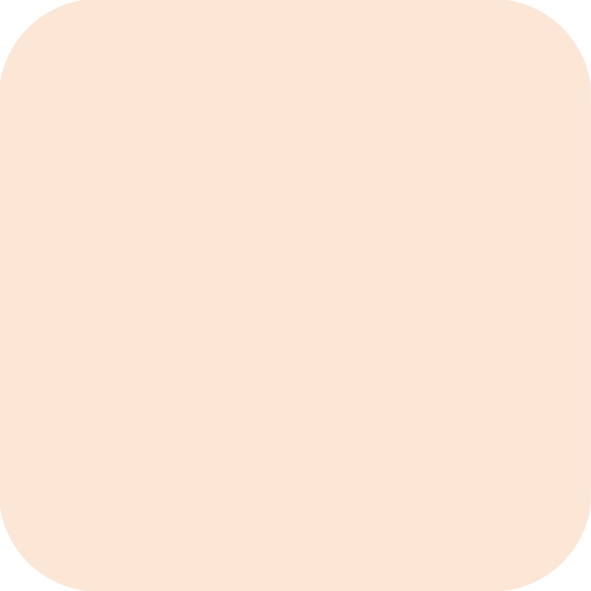}} exemplifies the historical context, while the darker orange \raisebox{-0.7ex}{\includegraphics[height=0.32cm]{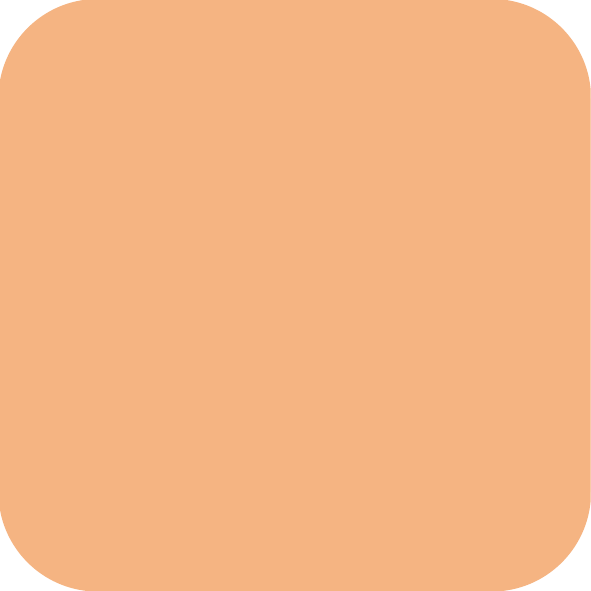}} exemplifies the future targets. The green blocks \raisebox{-0.7ex}{\includegraphics[height=0.32cm]{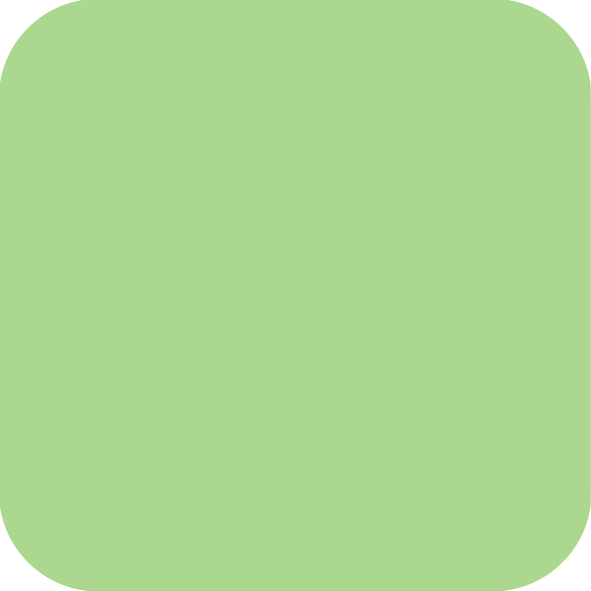}} denote the fine-grained interpolation actions.}
    \label{fig:strategy}
\end{figure}

As previously mentioned, our diffusion policy with coupled diffusion heads outputs a \emph{fixed-length} trajectory of $N$, where $N$ is determined by the sparse-action horizon $N_1$ and the interpolation horizon $N_2$, as illustrated in Fig.~\ref{fig:strategy}(a). And their relationship is given by:
\begin{equation}
    N = N_1 + (N_1 - 2) * N_2
\end{equation}

During training, we downsample the expert demonstration $\hat{\mathcal{A}}$ to generate sparse anchor labels $\hat{\mathcal{A}}^{\prime}$, which are used to supervise the \texttt{GDH}, as shown in Fig.~\ref{fig:strategy}(b). To enable closed-loop inference of \texttt{GDH} in dynamic environments, we further divide the sparse anchor labels into historical and future parts by randomly sampling a Gaussian-distributed index. Subsequently, as depicted in Fig.~\ref{fig:strategy}(c), the entire expert demonstration is segmented into $N_1$ contiguous segments $\{\hat{\mathcal{A}}_{i}\}_{i=1}^{N_1}$ using the downsampling indices. These segments serve as supervision labels for the \texttt{LDH}.

\subsection{Binary Mask Used in Trajectory-Steering Strategy}
\label{sec:binary_mask}
In this section, we formally define the three binary masks used in Eq.~\eqref{eq:trajectory-steering_GDH}, Eq.~\eqref{eq:trajectory-steering_LDH}, and Eq.~\eqref{eq:trajectory-steering_GDH_prime}:
\begin{itemize}
    \item \emph{Global Denoising Mask} \(\mathcal{M}_{\text{GDH}}\):  
    This mask with the length of \(N_1\) is used to inject the previously executed anchors $\mathcal{A}^{\prime}_{-}$ into the denoising process of \texttt{GDH} at current step $i$. All entries up to step \(i\) are set to \(1\), and the remaining entries are set to \(0\). \ie
    \begin{equation}
        \mathcal{M}_{\texttt{GDH}}[j] = 
        \begin{cases} 
            1, & j < i, \\
            0, & j \geq i.
        \end{cases}
    \end{equation}

    \item \emph{Local Denoising Mask} \(\mathcal{M}_{\text{LDH}}\):  
    This mask with the length of \(N_2\) is used to inject the neighboring anchor pair $(a_i^{\prime}, a_{i+1}^{\prime})$ into the denoising process of \texttt{LDH}  at current step $i$. Only the first and the last entries are set to \(1\), and all other entries are set to \(0\). \ie
    \begin{equation}
        \mathcal{M}_{\texttt{LDH}}[j] = 
        \begin{cases} 
            1, & j = 1 \ \text{or} \ j = N_2, \\
            0, & \text{otherwise}.
        \end{cases}
    \end{equation}

    \item \emph{Referring-Augmented Global Mask} \(\mathcal{M}^{\prime}_{\texttt{GDH}}\):  
    This mask with the length of \(N_1\) is used to inject the referring action $\mathcal{P}^{a}$ into the denoising process of referring-aware \texttt{GDH} at current step $i$. It extends \(\mathcal{M}_{\texttt{GDH}}\) by additionally setting the entry at index \(k\) to \(1\), while preserving all entries equal to \(1\) in \(\mathcal{M}_{\text{GDH}}\). Here \(k\) denotes the temporal position of the referring action \(\mathcal{P}^{a}\) as predicted by Eq.~\eqref{eq:TP}. \ie
    \begin{equation}
        \mathcal{M}^{\prime}_{\text{GDH}}[j] = 
        \begin{cases} 
            1, & j \leq i \ \text{or} \ j = k, \\
            0, & \text{otherwise}.
        \end{cases}
    \end{equation}
\end{itemize}

\subsection{Benchmark Modification For Evaluation}
\label{sec:benchmark_modification}
\begin{figure}[ht!]
    \centering
    \setlength{\fboxrule}{0pt}
    \framebox{{\includegraphics[width=0.6\linewidth]{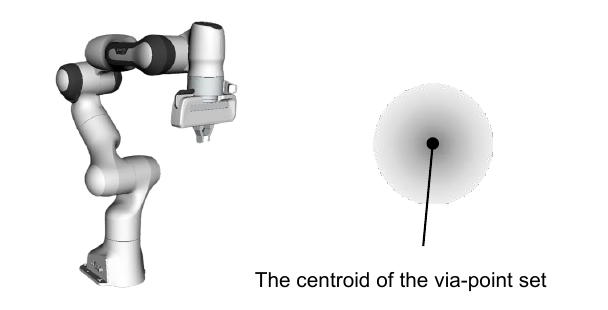}}}
   \caption{\textbf{Via-Points Generation.} Via-points are generated via Gaussian sampling around a centroid derived from the robot's initial configuration.}
    \label{fig:via-point}
\end{figure}
As discussed previously, we evaluate the referring-awareness capability of \mname{} by augmenting the involved simulation and real-world benchmarks with randomized via-points. These via-points are constrained to the robot's operational workspace to ensure reachability. The detailed procedure for this is outlined in Fig.~\ref{fig:via-point}: (1) defining a centroid for the via-point set based on the robot's initial configuration in each task, and (2) generating via-points via Gaussian sampling around this centroid for evaluation.

\section{Ablation Study}
\subsection{Ablation Study on Hyperparameters}
\label{sec:ablation_study_on_hyperparameters}
\begin{figure*}[h!]
  \centering
  \setlength{\fboxrule}{0pt}
  \begin{minipage}[b]{0.33\linewidth}
    \centering
    \begin{subfigure}{\linewidth}
      \framebox{\includegraphics[width=\linewidth]{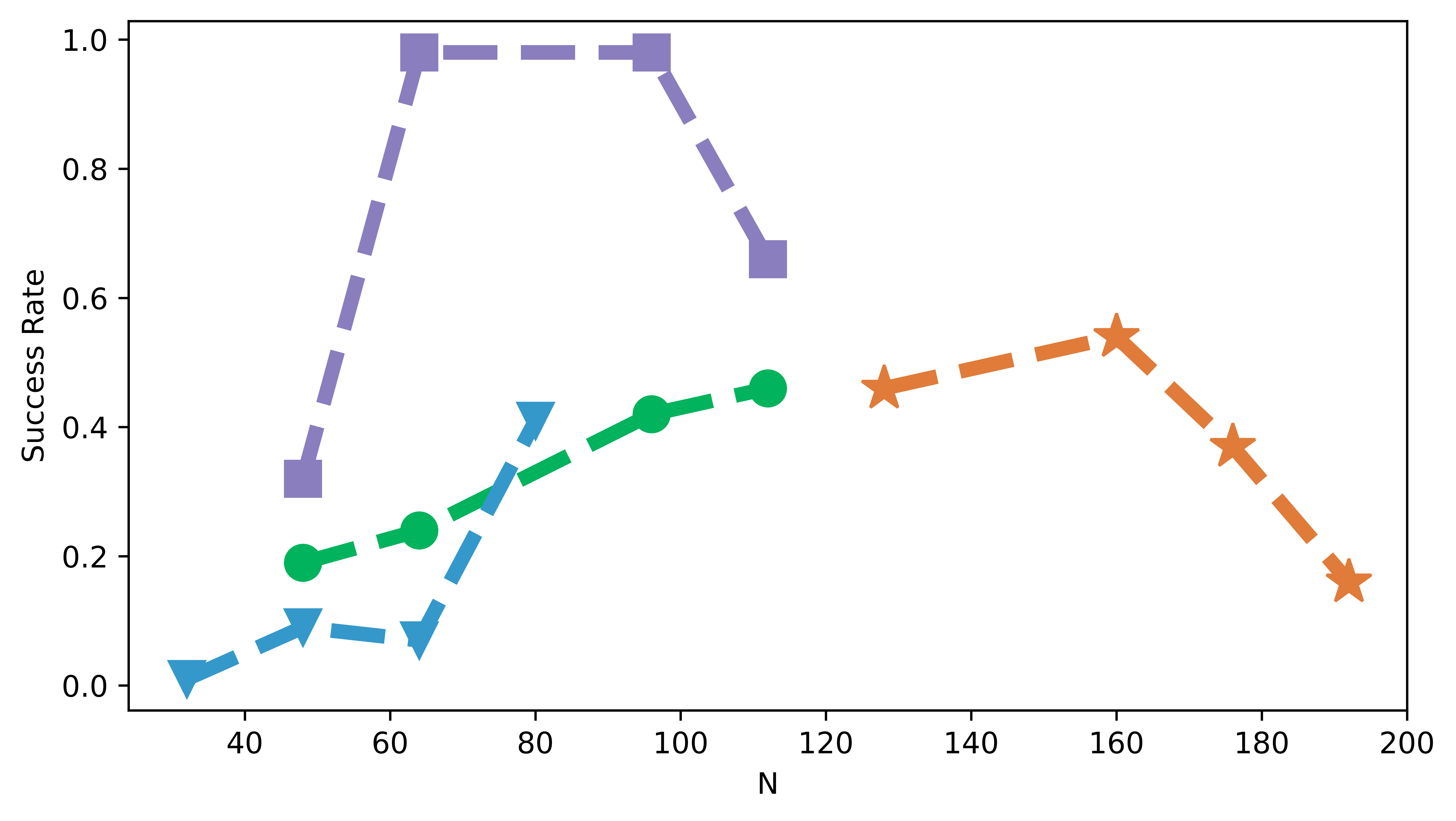}}
      \caption{Total Trajectory Length}
      \label{fig:ablation_trajectory_length}
    \end{subfigure}
  \end{minipage}%
  \hfill
  \begin{minipage}[b]{0.33\linewidth}
    \centering
    \begin{subfigure}{\linewidth}
      \framebox{\includegraphics[width=\linewidth]{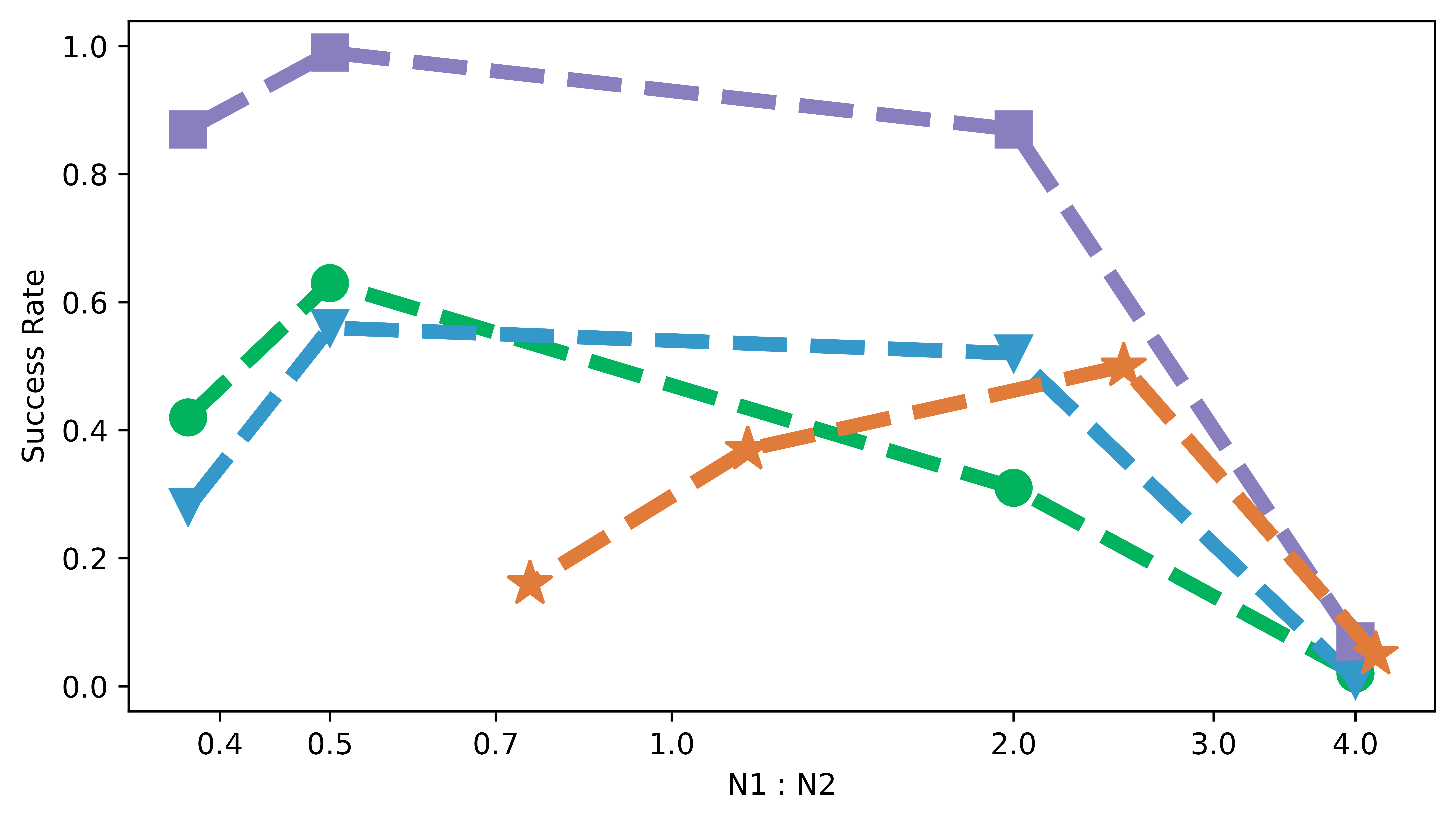}}
      \caption{Trajectory-Length Allocation}
      \label{fig:ablation_local_horizon}
    \end{subfigure}
  \end{minipage}
  \hfill
  \begin{minipage}[b]{0.33\linewidth}
    \centering
    \begin{subfigure}{\linewidth}
      \framebox{\includegraphics[width=\linewidth]{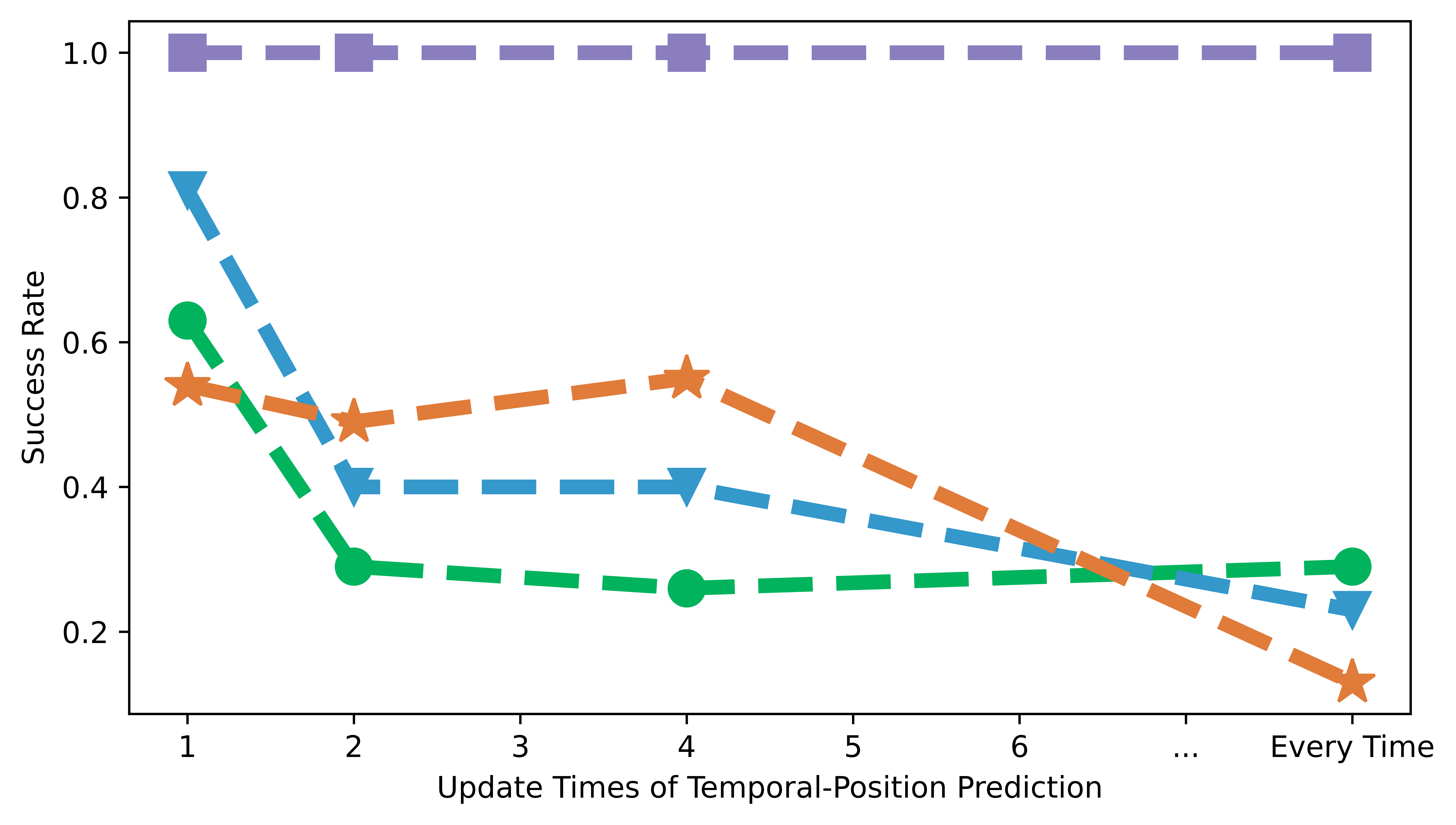}}
      \caption{Update Frequencies of T-P Prediction}
      \label{fig:ablation_update_times}
    \end{subfigure}
  \end{minipage}

  \vspace{4pt}
  \begin{minipage}{\linewidth}
    \centering
    \includegraphics[width=.5\linewidth]{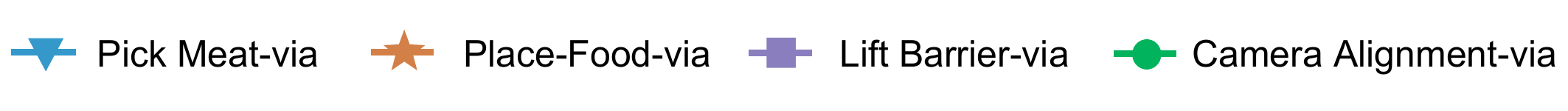}
  \end{minipage}
  
  \caption{\textbf{Visualization of ablation studies on key hyperparameters.}
    \textbf{(a)}~\emph{Total trajectory length}~$N$: performance peaks when $N$ matches task complexity, enabling the model to capture task-specific execution patterns.  
    \textbf{(b)}~\emph{Trajectory-length allocation} $N_1\!:\!N_2$: balanced ratios ($1\!:\!2 \sim 2\!:\!1$) allow our policy model to learn the entire trajectory most effectively.  
    \textbf{(c)}~\emph{Update Frequencies of T-P prediction}: excessive updates cause anchor drift and degrade performance. Here, success rate is defined as the fraction of trials that simultaneously achieve task completion and pass through the designated referring point, and all ablations were conducted under controlled variables.}
  \label{fig:ablation_total}
\end{figure*}
\noindent
\textbf{Total Trajectory Length.} 
As noted in Sec.~\ref{sec:closed-loop_global_policy}, our \mname{} generates \emph{fixed-length} trajectories of \(N\), which must be selected a priori for each task.  
To quantify sensitivity to this choice, we sweep \(N\) across tasks of increasing complexity, and the results are summarized in Fig.~\ref{fig:ablation_trajectory_length}.  
Quantitatively, an overly short \(N\) prevents our \mname{} from capturing the complete execution pattern required for each task, whereas an excessively long \(N\) injects redundant and ineffective information during training, degrading performance.

\noindent
\textbf{Trajectory-Length Allocation.}  
The total trajectory length $N$ is determined by the sparse-action horizon $N_1$ produced by \texttt{GDH} and the interpolation horizon $N_2$ inserted by \texttt{LDH}. We perform an ablation to quantify the sensitivity of \mname{} to the balance between these two components while keeping the overall length $N$ fixed for each task. As shown in Fig.~\ref{fig:ablation_local_horizon}, performance peaks around balanced ratios (\ie $1\!:\!2 \sim 2\!:\!1$), whereas larger imbalances consistently degrade results. We attribute this decline to two capacity-mismatch effects: 1) \emph{\texttt{GDH} overload.} A large $N_1\!:\!N_2$ ratio places the burden of modeling complex execution patterns almost entirely on \texttt{GDH}, making it difficult to fit the highly non-linear action manifold. 2) \emph{Supervision dilution.} When the trajectory is over-segmented (\ie $N_2$ is too small), the differences between adjacent segments fall below the noise floor.  Although \texttt{LDH} employs a learnable, temporal-dependent interpolation strategy, the regression targets become vanishingly simply and the network cannot learn meaningful interpolation kernels for each temporal position. 
% In short, \emph{GDH} and \emph{LDH} must operate in matched regimes: GDH provides ``sparse yet informative'' anchors, and \emph{LDH} completes the remaining complexity; either module starved or overloaded degrades the overall representation.

\noindent
\textbf{Temporal-Position Prediction.}
In the preceding experiments, the temporal-position prediction module is invoked only once at the beginning of inference. Although it could in principle be updated every inference step (Fig.~\ref{fig:temporal_position_prediction}), recomputing the temporal position for the same $\mathcal{P}$ repeatedly causes the anchor to drift, producing unstable trajectories.
To quantify this effect, we ablate this module by invoking it every $i$~steps ($i\in\{1,2,4,N\}$).  
Fig.~\ref{fig:ablation_update_times} shows that the more often we re-predict the position for an \emph{unchanged} $\mathcal{P}$, the larger the performance drop.
However, this does not imply that our model is unable to cope with moving or newly-appearing referring points.  
Whenever the referring point changes—either because objects move or because new ones appear—we simply reset the sparse anchor history $\mathcal{A}^{\prime}_{-}$ and restart trajectory generation from the current observation.  
In this way, \mname{} immediately adapts to the new configuration without suffering from anchor jitter.

\subsection{Experimental Setup for Ablating the Coupled Diffusion Heads}
\noindent
\textbf{Baselines.}  
Following the protocol in Sec.~\ref{exp:trajectory-steering}, we retain ACT, DP3 and CDP as baselines.  
In this experiment, we withhold $\mathcal{P}$ from all methods—including our \mname{}—in order to evaluate the intrinsic capability of our policy model with coupled diffusion heads against the myopic window-sliding paradigms employed by the baselines.
All involved methods are trained on an identical set of expert demonstrations for the same number of epochs, guaranteeing that any performance discrepancy is attributable solely to architectural factors.

\noindent
\textbf{Benchmarks.} Following \cite{Ze2024DP3,ma2025cdp,su2025dense}, we curate a cross-benchmark suite that spans Adroit~\cite{rajeswaran2017learning}, DexArt~\cite{bao2023dexart}, MetaWorld~\cite{yu2020meta}, and RoboFactory to evaluate the effectiveness of our policy model in various aspects: gripper-based and dexterous manipulation, articulated and rigid objects manipulation, and single- and multi-agent cooperation.

\subsection{Ablation Study on Infeasible Referring Points}
\label{sec:design_of_infeasible_referring_points}
As shown in Tab.~\ref{tab:infeasible_referring_points}, we introduce a set of deliberately infeasible referring points to rigorously evaluate whether \mname{} can faithfully adhere to the provided guidance even in these challenging scenarios. The definition of these infeasible referring points are as follows.

\noindent \textbf{Inside Camera.} In the Camera Alignment task, referring points are adversarially placed on the camera body itself. We uniformly sample 3D locations on the visible surface of the camera to test if the model blindly follows guidance that logically conflicts with the task objective.

\noindent \textbf{Inside Pot.} For the Place Food task, referring points are constrained inside the pot's volume. Points are randomly sampled within the pot's cylindrical cavity (excluding the bottom center to avoid trivial solutions), simulating an erroneous instruction to place food.

\noindent \textbf{Under Table.} In the Pick Meat task, referring points are hidden beneath the table. The points are randomly distributed within a rectangular region under the tabletop, creating a persistent occlusion that requires the model to reconcile the guidance with the impossibility of direct reaching.

\noindent \textbf{Out of Reach.} Also in the Pick Meat task, referring points are placed beyond the robot's workspace. Each point is fixed at 2 meters height above the table, with its (x, y) coordinates uniformly randomized within a 0.5m × 0.5m area centered above the workspace boundary, ensuring unambiguous physical infeasibility.

% \begin{table}[t]
%     \centering
%     \renewcommand{\arraystretch}{1.5}
%     \caption{\textbf{Ablation study on infeasible referring points.} Different from \texttt{SuR}, \texttt{SR} denotes the success rate regardless of penetrations.
%     }
%     \label{tab:infeasible_referring_points}
%     \resizebox{0.7\textwidth}{!}{
%     \begin{tabular}{c c c c c c c c c}
%         \toprule[1pt]
%         \multirow{2}{*}{Method} & \multicolumn{2}{c}{Inside Camera} & \multicolumn{2}{c}{Inside Pot} & \multicolumn{2}{c}{Under Table} & \multicolumn{2}{c}{Out of Reach} \\ \cmidrule(lr){2-3} \cmidrule(lr){4-5} \cmidrule(lr){6-7} \cmidrule(lr){8-9} 
%         & \texttt{RePR(}$\uparrow$\texttt{)} & \texttt{SR(}$\downarrow$\texttt{)} & \texttt{RePR(}$\uparrow$\texttt{)} & \texttt{SR(}$\downarrow$\texttt{)} &  \texttt{RePR(}$\uparrow$\texttt{)} & \texttt{SR(}$\downarrow$\texttt{)} &  \texttt{RePR(}$\uparrow$\texttt{)} & \texttt{SR(}$\downarrow$\texttt{)} \\
%         \hline 
%         DP3 & 0\% & 26\% & 0\%  & 15\% & \textbf{0\%} & 13\% & \textbf{0\%}  & 11\% \\  
%         CDP & 0\% & 84\% & 11\%  & 48\% & \textbf{0\%} & 82\% & \textbf{0\%}  & 86\% \\ 
%         \rowcolor{lightblue} 
%         \mname{} & \textbf{100\%} & \textbf{0\%} & \textbf{100\%}  & \textbf{0\%} & \textbf{0\%} &\textbf{0\%} & \textbf{0\%}  & \textbf{0\%} \\ 
%         \bottomrule[1pt]
%     \end{tabular}
%     }
%     \vspace{-\baselineskip}
% \end{table}

\begin{table}[t]
    \centering
    \renewcommand{\arraystretch}{1.5}
    \caption{\textbf{Ablation study on infeasible referring points.} 
    }
    \label{tab:infeasible_referring_points}
    \resizebox{0.5\textwidth}{!}{
    \begin{tabular}{c c c c c}
        \toprule[1pt]
        Method & Inside Camera & Inside Pot & Under Table & Out of Reach \\
        \hline 
        DP3 & 0\% & 0\% & \textbf{0\%} & \textbf{0\%} \\  
        CDP & 0\% & 11\% & \textbf{0\%} & \textbf{0\%} \\ 
        \rowcolor{lightblue} 
        \mname{} & \textbf{100\%} & \textbf{100\%} & \textbf{0\%} &\textbf{0\%} \\ 
        \bottomrule[1pt]
    \end{tabular}
    }
    \vspace{-\baselineskip}
\end{table}

\subsection{Design of OOD-yet-feasible Referring Points}
\label{sec:design_of_out-of-distribution_referring_points}
As illustrated in Fig.~\ref{fig:ood_yet_feasible}, the OOD-yet-feasible referring points are generated through the following two-step procedure:
\begin{figure}[ht!]
    \centering
    \setlength{\fboxrule}{0pt}
    \framebox{{\includegraphics[width=0.6\linewidth]{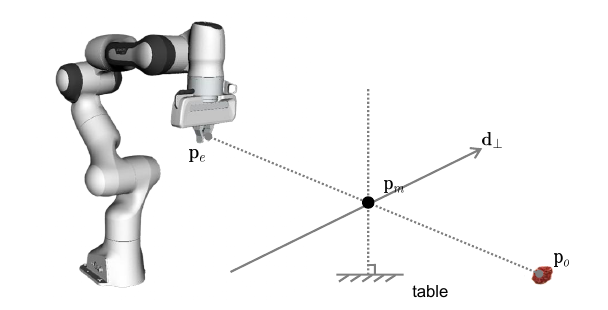}}}
   \caption{\textbf{OOD-yet-Feasible Referring Points Generation.}}
    \label{fig:ood_yet_feasible}
\end{figure}
\begin{enumerate}
    \item \textbf{Baseline Definition}: 
    Compute the midpoint $\mathbf{p}_m$ of the line segment connecting the robot end-effector's position $\mathbf{p}_e$ and the geometric center $\mathbf{p}_o$ of the target object:
    \[
    \mathbf{p}_m = \frac{\mathbf{p}_e + \mathbf{p}_o}{2}
    \]

    \item \textbf{Horizontal Sampling}: 
    On the plane parallel to the table surface, define a direction $\mathbf{d}_{\perp}$ that is perpendicular to the vector $\mathbf{p}_o - \mathbf{p}_e$ (\ie $\mathbf{d}_{\perp} \perp (\mathbf{p}_o - \mathbf{p}_e)$). 
    Then, generate a set of referring points:
    \[
    \left\{ \mathbf{p}_m + \lambda \cdot \mathbf{d}_{\perp} \mid \lambda \in \Lambda \right\}
    \]
    where $\Lambda$ is a predefined set of offsets that places the generated points outside the training data distribution while ensuring they remain kinematically reachable by the robot. And In this papaer, we set $\Lambda = \{0.1, 0.2, 0.3, 0.4\}$ m.
\end{enumerate}

\section{Simulation Experiments Details}
\subsection{Training Settings}
Each policy is trained independently on a single NVIDIA GeForce RTX 4090 GPU. We employ the AdamW optimizer with a learning rate of $1.0 \times 10^{-4}$, betas of $(0.95, 0.999)$, and $\epsilon = 1.0 \times 10^{-8}$. The learning rate undergoes a warmup phase for the first 500 steps, followed by training for the designated number of epochs specific to each benchmark task. The complete set of training parameters for all simulation experiments is provided in Tab.~\ref{tab:simulation-parameters}.
\begin{table}[ht!]
    \centering
    \caption{ \textbf{All Training Settings} for simulation experiments. }
    \resizebox{0.42\textwidth}{!}{
    \begin{tabular}{@{}lcc@{}} % 修改为居中对齐
        \toprule
        Benchmark & Parameter & Value \\ \midrule
        \multirow{5}{*}{Adroit} & Demonstrations Number & 10 \\
                                & Size of Point Clouds & (512, 3) \\
                                & Size of Images &  (84, 84, 3) \\
                                & Batch Size & 32 \\
                                & Epoch & 3000 \\ \midrule
        \multirow{5}{*}{Dexart} & Demonstrations Number & 100 \\
                                & Size of Point Clouds & (1024, 3) \\
                                & Size of Images &  (84, 84, 3) \\
                                & Batch Size & 32 \\
                                & Epoch & 3000 \\ \midrule
        \multirow{5}{*}{MetaWorld} & Demonstrations Number & 10 \\
                                & Size of Point Clouds & (512, 3) \\
                                & Size of Images &  (128, 128, 3) \\
                                & Batch Size & 32 \\
                                & Epoch & 3000 \\ \midrule
        \multirow{5}{*}{RoboFactory} & Demonstrations Number & 150 \\
                                & Size of Point Clouds & (512, 3) \\
                                & Size of Images &  (128, 128, 3) \\
                                & Batch Size & 128 \\
                                & Epoch & 500 \\
        \bottomrule
    \end{tabular}
    }
    \label{tab:simulation-parameters}
\end{table}

\subsection{Trajectory Length Allocation of our \mname{}}
For each task in our simulation and real-world experiments, we set the values of $N$, $N_1$, and $N_2$ based on its execution complexity (\cf Tab.~\ref{tab:trajectory-allocation}).
\begin{table}[ht!]
    \centering
    \caption{ \textbf{Trajectory-Length Allocation} for each task in simulation and real-world experiments.}
    \resizebox{0.45\textwidth}{!}{
    \begin{tabular}{@{}lcccc@{}}
        \toprule
        Benchmark & Task & \(N\) & \(N_1\) & \(N_2\)  \\ 
        \midrule
        \multirow{2}{*}{Adroit} & Door & 54 & 6 & 12 \\  
                                & Pen & 70 & 6 & 16 \\
        \midrule
        \multirow{3}{*}{Dexart} & Bucket & 11 & 3 & 8 \\
                                & Laptop & 20 & 4 & 8 \\
                                & Toilet & 70 & 6 & 16 \\
        \midrule
        \multirow{4}{*}{MetaWorld} & Shelf Place & 11 & 3 & 8 \\ 
                                   & Soccer & 158 & 14 & 12 \\
                                   & Sweep Into & 164 & 20 & 8 \\
                                   & Reach & 200 & 24 & 8 \\
        \midrule
        \multirow{4}{*}{RoboFactory} & Pick Meat & 65 & 9 & 8 \\
                                     & Lift Barrier & 74 & 10 & 8\\
                                     & Camera Alignment & 74 & 10 & 8 \\
                                     & Place Food & 164 & 20 & 8 \\
        \midrule
        \multirow{4}{*}{Real-World} & Collect Objects & 128 & 16 & 8 \\
                                     & Moving Playing Card Away & 128 & 16 & 8 \\
                                     & Grabbing Rod & 200 & 24 & 8  \\
        \bottomrule
    \end{tabular}
    }
    \label{tab:trajectory-allocation}
\end{table}

\subsection{Evaluation Metric}
We follow the evaluation protocol from DP3. For the Adroit, DexArt, and MetaWorld benchmarks, each experiment is run over three seeds (0, 1, 2). For each seed, the policy is evaluated over 20 episodes every 200 epochs, with the mean of the top-5 success rates recorded. The final performance is reported as the mean and standard deviation across the three seeds.
For the RoboFactory benchmark, each experiment is conducted with a single seed (0) and evaluated over 100 episodes at epoch 300.
This consistent protocol ensures a fair comparison between \mname{} and the baseline methods.

\subsection{Additional Qualitative Results}
\begin{figure*}[t]
    \centering
    \setlength{\fboxrule}{0pt}
    \framebox{{\includegraphics[width=\linewidth]{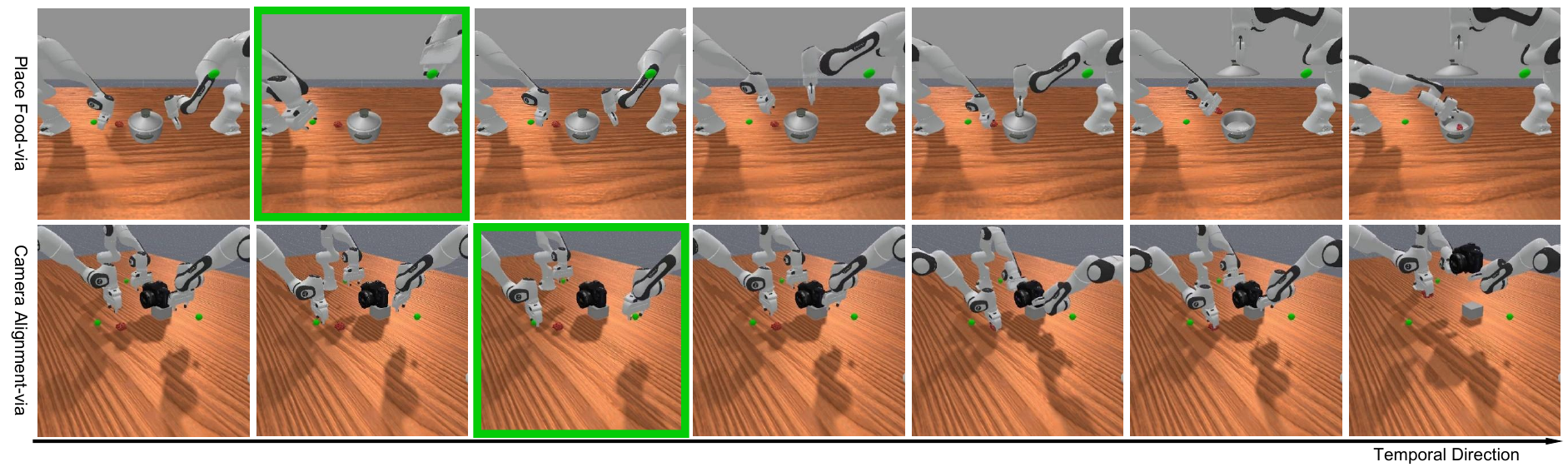}}}
    \caption{\textbf{Visualization of the trajectories generated by \mname{}} on \emph{Place-Food-via} and \emph{Camera-Alignment-via}. Here, we use green bounding boxes to mark the frames in which the end-effector passes through the designated via-point (green ball).}
    \label{fig:quality_results_2}
\end{figure*}

In this section, we visualize the additional qualitative results of the proposed \mname{} in Fig.~\ref{fig:quality_results_2}, which demonstrate the effectiveness of our model in referring-aware robotic manipulation.

%===============================================================================
\section{Real-world Experiments Details}
\label{sec:real-world_experiments_details}
\subsection{Platform} 
\begin{figure}[ht!]
    \centering
    \setlength{\fboxrule}{0pt}
    \framebox{{\includegraphics[width=0.8\linewidth]{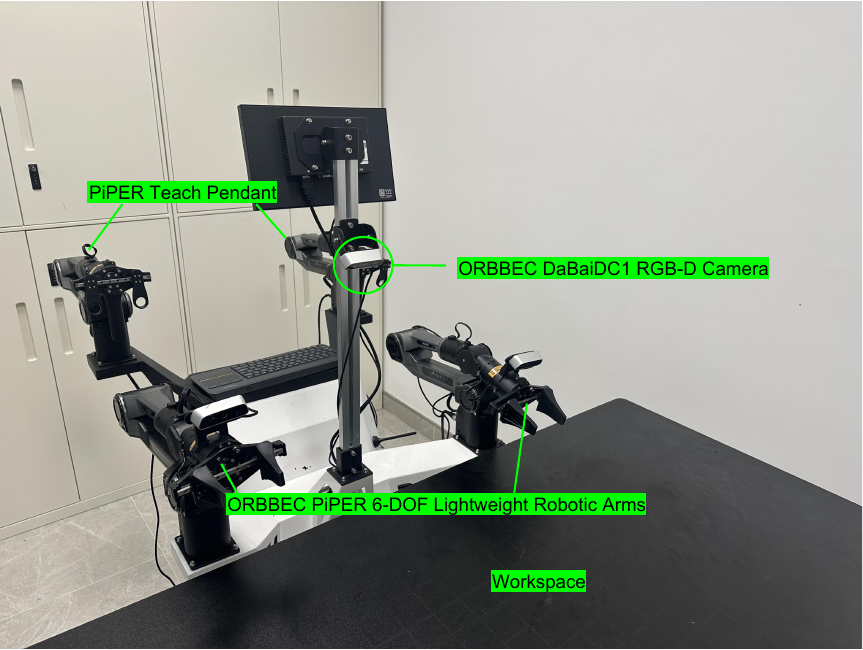}}}
    \caption{\textbf{Real-world Experimental Platform.} The setup comprises a dual-robot arm system, a top-down RGB-D camera, and a black table serving as the workspace.}
    \label{fig:workspace}
\end{figure}

As illustrated in Fig.~\ref{fig:workspace}, we conducted the real-world experiments with a dual-arm setup composed of two ORBBEC PiPER 6-DOF lightweight manipulators, each fitted with a two-finger gripper. An externally mounted, top-down ORBBEC DaBaiDC1 RGB-D sensor delivered a global view of the workspace. Demonstrations were collected through two factory PiPER Teach Pendants that allow simultaneous teleoperation of both arms. A single workstation (NVIDIA GeForce RTX 4090) handles the entire data pipeline: recording observations, performing policy inference and streaming commands to the arm controllers at 30 Hz.

\subsection{Tasks}
We construct out real-world modified benchmark based on five original tasks: \emph{Collecting Objects}, \emph{Push T}, \emph{Stacking Playing Card}, \emph{Grabbing Rod} and \emph{Handing Eraser Over}. The specific description for each original task is provided in Tab.~\ref{tab:real-tasks}. 
Following this, we inject a mandatory via-point into these original tasks to assess the referring-awareness of all involved methods.  
The resulting tasks are termed \emph{Collecting Objects-via}, \emph{Push T-via}, \emph{Stacking Playing Card-via}, \emph{Grabbing Rod-via} and \emph{Handing Eraser Over-via}.
\begin{table}[ht!]
    \centering
    \caption{\textbf{Original Task Descriptions} for real-world experiments.}
    \resizebox{\textwidth}{!}{
    \begin{tabular}{p{4cm} c p{11cm}}
        \toprule
        Task & Agent Number & Description \\  \midrule
        \emph{Collecting Objects} & 1 & A doll is placed on the table. The robotic manipulator first grasps it and then transports it into the green box. \\ \midrule
        \emph{Push T} & 1 & A T-shaped object is initially positioned on the table. The robotic arm pushes it into a predefined T-shaped location. \\ \midrule
        \emph{Stacking Playing Card} & 1 & A playing card is placed flat in the central region of the table. The robotic arm first grasps it precisely and then places it vertically into another playing card to achieve a stacking effect. \\ \midrule
        \emph{Grabbing Rod} & 2 & A long rod is placed on a block. The two robotic arms first simultaneously grasp each end of the rod and then collaboratively lift it to a specified height. \\ \midrule
        \emph{Handing Eraser Over} & 2 & One robotic arm first grasps an eraser precisely. It then passes the eraser to another robotic arm through a coordinated handover motion. \\
        \bottomrule
    \end{tabular}
    }
    \label{tab:real-tasks}
\end{table}

\subsection{Demonstrations.} 
The demonstrations utilized in our real-world experiments were generated by teleoperating the dual-arm system with the PiPER Teach Pendants. For each task, we collected a total of 50 demonstrations, each carefully selected to explicitly exhibit the salient motion and object contacts required for reliable success; episodes that deviated from these criteria were discarded and re-recorded.

% \subsection{Training Settings}
% \begin{table}[ht!]
%     \centering
%     \caption{ \textbf{All Training Settings} for real-world experiments. }
%     \resizebox{0.32\textwidth}{!}{
%     \begin{tabular}{@{}lc@{}} % 修改为居中对齐
%         \toprule
%         Parameter & Value \\ \midrule
%         Demonstrations Number & 50 \\
%         Size of Images &  (256, 256, 3) \\
%         Batch Size & 32 \\
%         Epoch & 500 \\
%         \bottomrule
%     \end{tabular}
%     }
%     \label{tab:simulation-parameters}
% \end{table}

%%%%%%%%%%%%%%%%%%%%%%%%%%%%%%%%%%%%%%%%%%%%%%%%%%%%%%%%%%%%%%%%%%%%%%%%%%%%%%%
%%%%%%%%%%%%%%%%%%%%%%%%%%%%%%%%%%%%%%%%%%%%%%%%%%%%%%%%%%%%%%%%%%%%%%%%%%%%%%%

\end{document}